\documentclass[sigconf]{acmart}

\usepackage{adjustbox}
\usepackage{multirow}

\usepackage{colortbl} 
\usepackage{bbding} 
\usepackage{pifont} 
\usepackage{balance} 

\AtBeginDocument{%
  }

\copyrightyear{2025}
\acmYear{2025}
\setcopyright{acmlicensed}\acmConference[MM '25]{Proceedings of the 33rd ACM International Conference on Multimedia}{October 27--31, 2025}{Dublin, Ireland}
\acmBooktitle{Proceedings of the 33rd ACM International Conference on Multimedia (MM '25), October 27--31, 2025, Dublin, Ireland}
\acmDOI{10.1145/3746027.3755282}
\acmISBN{979-8-4007-2035-2/2025/10}

\acmSubmissionID{2977}
\begin{document}

\title{Explicit Context Reasoning with Supervision for Visual Tracking}

\author{Fansheng Zeng}
\affiliation{%
  \institution{Guangxi Normal University}
  \city{Guilin}
  \country{China}
}
\email{zengfansheng@stu.gxnu.edu.cn}

\author{Bineng Zhong}
\affiliation{%
  \institution{Guangxi Normal University}
  \city{Guilin}
  \country{China}
}
\authornote{Corresponding Author.}
\email{bnzhong@gxnu.edu.cn}

\author{Haiying Xia}
\affiliation{%
  \institution{Guangxi Normal University}
  \city{Guilin}
  \country{China}
}
\email{xhy22@mailbox.gxnu.edu.cn}

\author{Yufei Tan}
\affiliation{%
  \institution{Guangxi Normal University}
 \city{Guilin}
 \country{China}
}
\email{jeffrey.yf.tan@gxnu.edu.cn}

\author{Xiantao Hu}
\affiliation{%
  \institution{Nanjing University of Science and Technology}
  \city{Nanjin}
  \country{China}
}
\email{xiantaohu@njust.edu.cn}

\author{Liangtao Shi}
\affiliation{%
  \institution{Hefei University of Technology}
  \city{Hefei}
  \country{China}}
\email{shilt@mail.hfut.edu.cn}

\author{Shuxiang Song}
\affiliation{%
  \institution{Guangxi Normal University}
  \city{Guilin}
  \country{China}}
\email{songshuxiang@mailbox.gxnu.edu.cn}
\authornotemark[1]

\renewcommand{\shortauthors}{Fansheng Zeng et al.}

\begin{abstract}
Contextual reasoning with constraints is crucial for enhancing temporal consistency in cross-frame modeling for visual tracking. However, mainstream tracking algorithms typically associate context by merely stacking historical information without explicitly supervising the association process, making it difficult to effectively model the target's evolving dynamics.
To alleviate this problem, we propose RSTrack, which explicitly models and supervises context reasoning via three core mechanisms.
\textit{1) Context Reasoning Mechanism}: Constructs a target state reasoning pipeline, converting unconstrained contextual associations into a temporal reasoning process that predicts the current representation based on historical target states, thereby enhancing temporal consistency.
\textit{2) Forward Supervision Strategy}: Utilizes true target features as anchors to constrain the reasoning pipeline, guiding the predicted output toward the true target distribution and suppressing drift in the context reasoning process.
\textit{3) Efficient State Modeling}: Employs a compression-reconstruction mechanism to extract the core features of the target, removing redundant information across frames and preventing ineffective contextual associations.
These three mechanisms collaborate to effectively alleviate the issue of contextual association divergence in traditional temporal modeling. Experimental results show that RSTrack achieves state-of-the-art performance on multiple benchmark datasets while maintaining real-time running speeds. 
Our code is available at \url{https://github.com/GXNU-ZhongLab/RSTrack}.
\end{abstract}




\maketitle
\section{Introduction}
\label{sec:intro}
\begin{figure}[t]
    \centering
        \includegraphics[width=\linewidth]{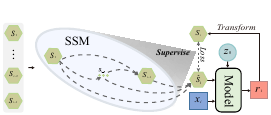}
    \caption{Our framework. The historical state tokens \( S_1 \) to \( S_{i-1} \) are fed into an state-space-model (SSM)-based reasoning module to generate the current predicted state \( \hat{S}_i \). This predicted state \( \hat{S}_i \), together with the initial template \( Z_0 \), is used to model the current search frame \( x_i \), resulting in the tracking output \( r_i \). After tracking, the true state \( S_i \) is extracted and used to supervise the reasoning process.}
    \label{fig_Paradigm_comparison}
\end{figure}

Visual object tracking \cite{siamfc, siamban, APTrack, uscltrack, SGLATrack, MAVLT} is a fundamental task in computer vision. 
It aims to locate the object in subsequent frames using a bounding box, given the initial state of an arbitrary object in the first frame.
However, in some complex tracking scenarios, such as fast motion and similar object interference or occlusion, traditional methods \cite{siamfc++, siamrpn, sbt} relying on a static initial frame often struggle to handle dynamic target appearance variations or background interference, leading to poor robustness.

To overcome the limitations of static references, researchers \cite{stark, artrack, STTrack, odtrack} have explored leveraging historical target features as priors to capture correlations with the current frame, thereby enabling dynamic modeling of both the target and the scene.
This topic has attracted widespread attention and developed rapidly, with mainstream methods transitioning from early discrete dynamic template updating \cite{stark,mixformer,stmtrack} to current continuous video-level context modeling \cite{artrackv2, aqatrack, mambalct}.
Overall, current video-level context modeling methods can be broadly categorized into two types:
\textit{1) Contextual Propagation}: 
This approach \cite{odtrack,evptrack} passes lightweight features or tokens along the temporal dimension to establish temporal association between the initial and current frames. However, the methods that aggregate target information from multiple frames using unified features or tokens often overlook the distinctive features of the target in different frames, making it hard to capture the contextual association accurately and resulting in poor temporal modeling.
\textit{2) Contextual Association Reasoning}: 
This method \cite{temtrack, aqatrack} generates a unique target state token for each frame. The state tokens from multiple historical frames form a continuous sequence of target states, which is then modeled and inferred along the temporal dimension. However, due to the lack of effective supervision during this process, the model struggles to fully capture the temporal dynamics of the target state, resulting in modeling outputs that deviate significantly from the actual target state.

To effectively learn and explicitly supervise the context reasoning process, we propose a novel context learning framework, RSTrack, aiming to model target consistency in video sequences to assist current state reasoning.
As shown in Fig.\ref{fig_Paradigm_comparison}, we innovatively use the true target state as a forward supervision signal to guide the continuous reasoning from historical to current states.
Compared with existing implicit context modeling approaches \cite{odtrack, aqatrack, mambalct}, RSTrack explicitly learns and leverages the temporal consistency among historical states to infer the current target state, thereby achieving context reasoning modeling with a clear evolutionary direction.
Specifically, RSTrack achieves the above objectives through the following three components:
\textbf{\textit{1) Efficient State Modeling}}: 
We propose a spatial-channel compression module that reduces redundant target features across frames by compressing them into state tokens that only retain essential object information. 
To ensure effective compression, we reconstruct the original features by combining the tokens with template features, and regularize the process using an L2 loss.
This mechanism not only enables efficient information storage but also mitigates interference caused by inter-frame redundancy during context reasoning, thereby avoiding invalid contextual associations.
\textbf{\textit{2) Context Reasoning Mechanism}}: 
We use a state space model to capture temporal variations of the target across frames, modeling the temporal correlations between historical target states to predict the current state. 
This predicted state token is converted to predicted target features via the reconstruction mechanism, which further refines the search feature in the temporal decoder. 
This mechanism fully leverages temporal consistency, enabling the model to reason more accurately about the current frame based on past states.
\textbf{\textit{3) Forward Supervision Strategy}}: To effectively constrain the context reasoning process, we compress the target features extracted from the visual encoder into compact state tokens, which serve as reliable supervision signals. By computing the L2 norm distance between the predicted state tokens and the supervision signals, we establish an explicit constraint mechanism that ensures effective learning of the contextual reasoning process and suppresses drift during the modeling process.
In summary, we make the following contributions:
\begin{itemize}
\item[$\bullet$] 
We propose a novel tracking framework named RSTrack. This framework explicitly supervises and models the temporal consistency between historical states, enabling robust cross-frame target state inference.
\item[$\bullet$] 
We design a spatial-channel compression module that retains core target information in each frame. This reduces computational costs and enables a more efficient contextual reasoning process.
\item[$\bullet$]
Our approach achieves a new state-of-the-art on multiple benchmarks, including LaSOT, \(\text{LaSOT}_{ext}\), GOT-10K, TrackingNet, TNL2K and UAV123.
\end{itemize}
\section{Related Works}
\subsection{Temporal modeling for visual tracking}
Traditional visual tracking methods \cite{siamfc, GMMT, MCTrack, FFTrack, DFAT} usually rely on the initial template frame as a reference to track the target in subsequent search frames. 
This approach, which depends solely on the initial static appearance, often performs poorly when the target undergoes significant appearance variations or similar object interference.
To address the problem, researchers \cite{stmtrack, stark, seqtrack} have increasingly begun to introduce additional target references. 
For example, STARK \cite{stark} introduces dynamic templates and updates them using an additional score head; 
MixFormer \cite{mixformer} updates templates through fixed intervals and a score prediction module; 
STMTrack \cite{stmtrack} stores historical target information within a memory network and utilizes the current frame as a query to retrieve highly relevant target features from the memory bank. 
However, these discrete accumulation methods generally perform poorly when modeling the continuous variations of the target.
More recently, many researchers \cite{odtrack, evptrack, mambalct} have focused on context propagation, using lightweight features or tokens to model cross-frame context. For example, ODTrack \cite{odtrack} propagates historical tokens to establish cross-frame associations; EVPTrack \cite{evptrack} abstracts multi-scale features into explicit visual prompts to enhance contextual connections. 
Nevertheless, these methods often overlook the uniqueness of features in different frames, as they aggregate historical information using unified features. 
Recent research, TemTrack \cite{temtrack}, generates unique target state tokens for each frame and models the sequence of historical frame state tokens along the temporal dimension. This approach preserves the uniqueness of the target in each frame, making the context modeling process more refined. 
However, without effective supervision, such implicit context learning may lead to model reasoning drift and fail to precisely capture the temporal cross-frame consistency of targets.
In contrast, our approach introduces a forward supervision mechanism that explicitly models the dynamic changes in the target appearance across the video sequence, effectively suppressing drift in the reasoning process and ensuring better temporal consistency.

\begin{figure*}[t]
    \centering
    \includegraphics[width=0.9\textwidth]{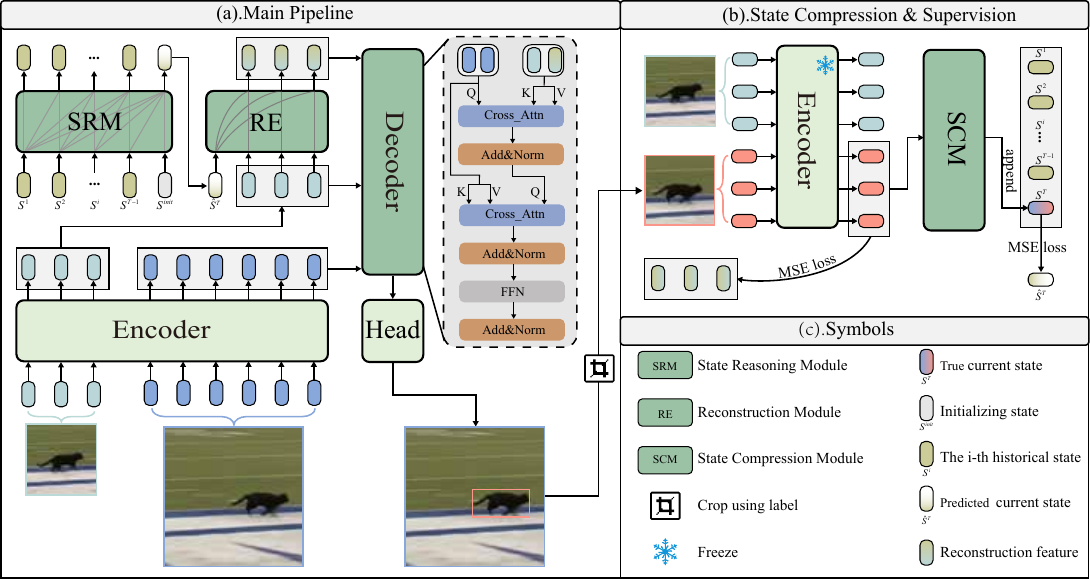}
    \caption{
    The architecture of RSTrack. 
    (a) Main Pipeline: The encoder extracts visual features from video frames. The state reasoning and reconstruction modules infer and reconstruct the predicted features, which the temporal decoder integrates to enhance visual representation. The prediction head then performs target localization.
    (b) State Compression and Supervision: True target features from the search region are extracted and compressed. Both pre- and post-compression features supervise the state reasoning process. The compressed state is appended to the historical sequence to update the target state.
    }
    \label{fig_Overall_framework}
\end{figure*}

\subsection{State Space Model for vision}
The State Space Model (SSM) \cite{ssm1,ssm2,ssm3} is originally developed for natural language processing, demonstrating exceptional capability in capturing dynamic changes and complex dependencies in sequences. 
With the introduction of Mamba \cite{mamba}, which integrates data-dependent SSM layers and an efficient parallel scanning selection mechanism, 
SSM has made significant progress in natural language processing and gained widespread attention in computer vision.
In visual classification tasks, VMamba \cite{vmamba} is the first to integrate Mamba into the design of visual backbone networks. By employing an innovative 2D selective scanning mechanism, it successfully extended Mamba's capabilities to image processing, providing a novel modeling paradigm for visual tasks. 
Subsequently, ViM \cite{vim} combined a bidirectional SSM mechanism to achieve data-dependent global visual context modeling, significantly reducing parameter complexity while achieving performance comparable to ViT \cite{vit}.
In medical video object segmentation tasks, ViViM \cite{vivim} effectively compresses long-term spatiotemporal representations into multi-scale sequences through a specially designed spatiotemporal Mamba block, thereby more accurately capturing key cues in video frames.
Additionally, in visual tracking tasks, MambaLCT \cite{mambalct} efficiently refines search features into temporal tokens via the Mamba mechanism,
utilizing them to transmit historical state information.
Meanwhile, TemTrack \cite{temtrack} combines Mamba’s long-sequence modeling ability with the global perception of the attention mechanism, thereby enhancing contextual understanding and adjusting the target state.
In this study, we leverage the advantages of Mamba in handling long-term dependencies and data-aware reasoning, applying it to the contextual reasoning process of target states, thereby achieving more accurate and robust target state prediction.

\section{Method}
In this section, we will provide a detailed explanation of the proposed RSTrack.
First, we present a brief overview of the overall framework.
Then, we describe each module in detail to explain its roles and interactions.
Finally, we introduce the training and inference pipelines to outline the complete workflow.

\subsection{Overview}
\label{subsec: Overview}
The overall framework of RSTrack is shown in Fig.\ref{fig_Overall_framework}. First, the template frame \(\mathcal{I}_{z}\) and search frame \(\mathcal{I}_{x}\) are embedded as 1D tokens and processed by a visual encoder to extract initial target features \(\mathcal{F}_{x}\). Subsequently, the state reasoning module uses the historical state sequence \(\Psi\) to predict the current target state. Based on this prediction, the reconstruction module transforms the template features \(\mathcal{F}_{z}\) to generate predicted target features \(\widehat{\mathcal{F}}_{tg}^t\).  
Next, the temporal decoder refines the initial search features \(\mathcal{F}_{x}\) using \(\widehat{\mathcal{F}}_{tg}^t\), producing the final temporally enriched features for the head network to locate the target. After each iteration, the true target features \(\mathcal{F}_{tg}^t\) are compressed and appended to \(\Psi\), maintaining the continuity of the historical states.
Additionally, these true features \(\mathcal{F}_{tg}^t\), along with the compressed state tokens, serve as supervisory signals to guide reasoning and explicitly capture contextual dynamics.

\begin{figure*}[t]
    \centering
        \includegraphics[width=\linewidth]{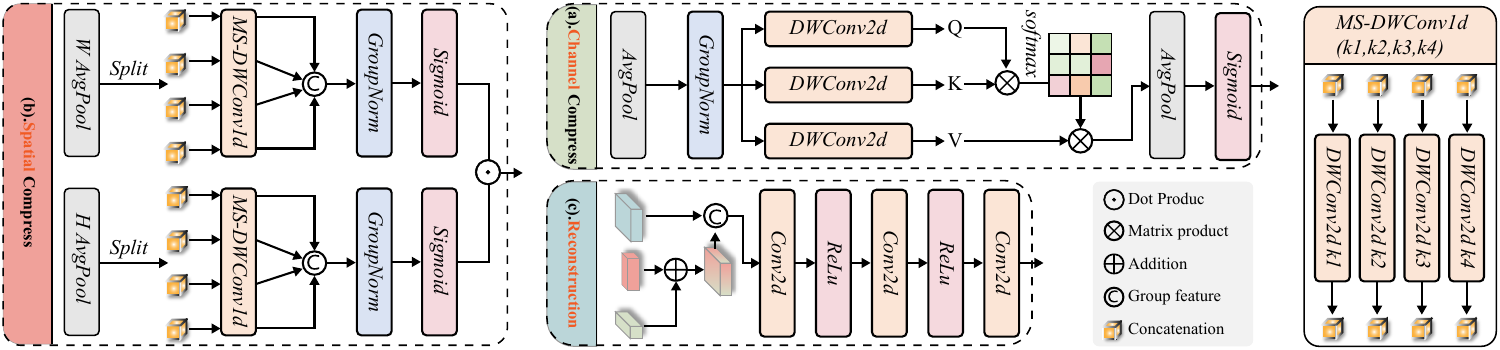}
    \caption{
    Compression-Reconstruction Mechanism. The detailed structures of the channel compression module and spatial compression module are shown in (a) and (b), respectively, while (c) presents the specific design of the reconstruction module.
    }
    \label{fig_Compression_Reconstruction}
\end{figure*}

\subsection{Visual Encoder}
\label{subsec: feature extraction network}
We adopt Fast-iTPN \cite{Fast-iTPN} as the visual encoder. 
Unlike ViT \cite{vit}, which directly embeds the image into patches of size 16×16, Fast-iTPN \cite{Fast-iTPN} utilizes a downsampling convolutional layer with a stride of 4 and two downsampling convolutional layers with a stride of 2 to progressively establish connections between adjacent patches. 
The main process of the visual encoder is as follows: First, the template frame 
$\mathcal{I}_{z} \in \mathbb{R}^{3 \times H_{z} \times W_{z}}$ 
and the search frame $\mathcal{I}_{x} \in \mathbb{R}^{3 \times H_{x} \times W_{x}}$
are processed through the aforementioned image embedding operation, 
resulting in the embedded features $\mathcal{P}_{z} \in \mathbb{R}^{D \times \frac{H_{z}}{16} \times \frac{W_{z}}{16}}$
and $\mathcal{P}_{x}\in \mathbb{R}^{D \times \frac{H_{x}}{16} \times \frac{W_{x}}{16}}$.
Then, $\mathcal{P}_{z}$ and $\mathcal{P}_{x}$ are flattened along the spatial dimension and concatenated to form
$\mathcal{F}_{v}^{0} = [\mathcal{P}_{z};\mathcal{P}_{x}] \in \mathbb{R}^{D \times (N_z+N_x)}$. 
Here, $N_z=H_{z}W_{z}/{16^2},\ N_x=H_{x}W_{x}/{16^2},\ D=512$. Next, $\mathcal{F}_{v}^{0}$ is fed into the visual encoder to enable the interaction and extraction of search features and template features. 
The process of the \(l^{th}\) visual encoder layer can be described by the following equation:
\begin{equation}
\begin{split}
&\mathcal{F}_{v}^{l} = \mathbb{E}^l(\mathcal{F}_{v}^{l-1}),\ l=1,2,...,L \\
\end{split}
\end{equation}
Here, the encoder layer comprises a multi-head self-attention mechanism and an MLP. See Fast-iTPN \cite{ Fast-iTPN} for details.

\subsection{Context Reasoning Mechanism}
\label{subsec: context evolution reasoning mechanism}
The context reasoning mechanism involves two phases: \textit{1)} predicting current target features via state reasoning; \textit{2)} refining target features by associating the reasoning results with search features.
To address inter-frame redundancy and high computational cost from modeling complete features, we introduce a feature compression-reconstruction mechanism in the state reasoning module for efficient video-level association.
We first describe this mechanism, then detail the state reasoning module and temporal decoder.

\subsubsection{\textbf{Compression-Reconstruction Mechanism.}} 
The compression process transforms complete target features into compact state tokens containing key information, while the reconstruction process restores the features by fusing the template $\mathcal{F}_{z}$ with these tokens.
Unlike simply training two inverse networks, we embed the reconstruction module into the state reasoning model, allowing it to use the predicted target state tokens \(\{\widehat{\mathcal{S}}^{T}, \widehat{\mathcal{C}}^{T}\}\) to reconstruct the target features \(\widehat{\mathcal{F}}_{tg}^t\).  
This design enables the state reasoning model to focus on learning the temporal variations of the target state, while the reconstruction network is dedicated to restoring the target features, effectively decoupling the target feature reasoning process. Through this mechanism, we reduce redundant contextual associations and also lower the computational cost.

\textbf{State Compression Module.}
The compression module consists of spatial compression and channel compression, with its detailed architecture illustrated in Fig.\ref{fig_Compression_Reconstruction}. 
\textit{\textbf{In the channel compression part}}, inspired by the powerful modeling capability of multi-head self-attention in ViT \cite{vit}, we propose a self-attention-based channel focusing mechanism. Unlike ViT \cite{vit}, we define the query, key, and value as \( Q, K, V \in \mathbb{R}^{B \times C \times N} \), meaning that self-attention is computed along the channel dimension.  
Furthermore, to reduce the computational cost, our compression method is based on average pooling, with the specific implementation details as follows:
\begin{equation}
\begin{split}
&\mathcal{C}_x =\varpi(\varrho_{1}(\mathcal{F}_{tg}^t)),\\
 &Q = \vartheta_{q}(\mathcal{C}_x), \quad
 K = \vartheta_{k}(\mathcal{C}_x), \quad
 V = \vartheta_{v}(\mathcal{C}_x); \\
&\mathcal{C}^{t} = \sigma(\varrho_{2}(\text{Attn}(Q,K,V))).
\end{split}
\end{equation}
Here, $\varrho_{1}$ denotes a $4 \times 4$ average pooling, $\varpi$ refers to Group Normalization with group size 1, and $\varrho_{2}$ is global average pooling. $\vartheta_{q}$, $\vartheta_{k}$, and $\vartheta_{v}$ are $1 \times 1$ depthwise separable convolutions, and $\sigma$ denotes the Sigmoid activation function.
\textit{\textbf{In the spatial compression part}}, we decompose the width and height dimensions of the features and assign independent sub-features to each dimension, fully capturing their spatial information. The process is as follows:
\begin{equation}
\begin{split}
&\mathcal{S}_{u} =\varrho_{u}(\mathcal{F}_{tg}^t), \quad \mathcal{S}_{u}^i = \mathcal{S}_{u}[:,(i-1)\times \frac{C}{K}:i\times \frac{C}{K},:],\\
&\tilde{\mathcal{S}}_{u} = Concat(\bigcup_{i=1}^{G} \vartheta^{i}_{u}(\mathcal{S}_{u}^i)), \quad i=1,2,...,G\\
&\mathcal{S}^{t} = \sigma(\varpi_u(\tilde{\mathcal{S}}_{w})) \times\sigma(\varpi_u(\tilde{\mathcal{S}}_{h})), \quad u \in [w, h]\\
\end{split}
\end{equation}
Here, \(\varrho_{u}\) denotes global pooling along the width or height dimension, \(\vartheta^{i}_{u}\) represents depthwise separable 1D convolution with a kernel size of \(1 \times 1\), \(Concat(\bigcup_{i=1}^{G}(x^i))\) indicates concatenating all \(x^i\) along the channel dimension, \(\varpi_u\) represents Group Normalization with a group size of 4, and \(\sigma\) is the Sigmoid activation function.
By accumulating compression results over multiple time steps, we can construct a history state sequence of the target. denoted as
\(
\Psi : [\{\mathcal{S}^{0}, \mathcal{C}^{0}\}, 
\{\mathcal{S}^{1}, \mathcal{C}^{1}\}, \dots,
\{\mathcal{S}^{t-1}, \mathcal{C}^{t-1}\}] 
\Leftarrow \{\mathcal{S}^{t}, \mathcal{C}^{t}\}.
\)
At each time step, $\Psi$ is updated by appending the current state tokens $\{\mathcal{S}^{t}, \mathcal{C}^{t}\}$ to the end, maintaining the target’s temporal states.

\textbf{Reconstruction Module.}
The reconstruction module generates the target predicted feature \(\widehat{\mathcal{F}}_{tg}^t\) by utilizing the predicted state tokens \(\{\mathcal{S}^{t}, \mathcal{C}^{t}\}\) and the template feature \(\mathcal{F}_{z}\).
The detailed structure is shown in Fig.\ref{fig_Compression_Reconstruction}(c). Specifically, the spatial state \(\widehat{\mathcal{S}}^{T}\) and the channel state \(\widehat{\mathcal{C}}^{T}\) are summed and broadcasted to match the shape of the template feature \(\mathcal{F}_{z}\). Then, the two are concatenated along the channel dimension. Subsequently, the concatenated features are passed through a series of multi-scale convolutional layers and activation layers to progressively integrate information from both sources, jointly constructing the target predicted feature \(\widehat{\mathcal{F}}_{tg}^t\).

\subsubsection{\textbf{State Reasoning Module.}} 
\begin{figure}[t]
    \centering
        \includegraphics[width=0.9\linewidth]{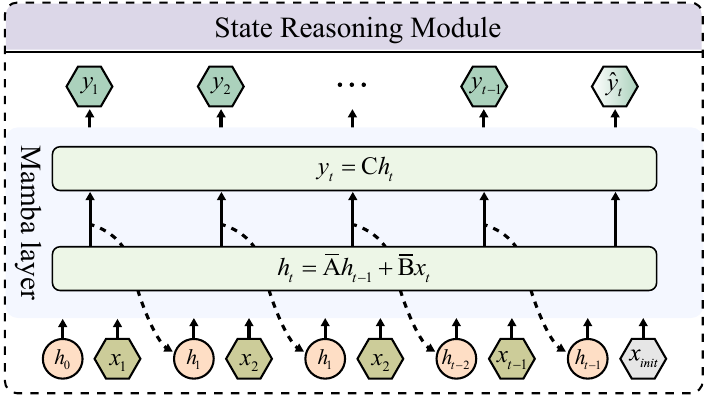}
    \caption{
    State Reasoning Module. The current state is first initialized by \( x_{init} \), then progressively inferred based on the dynamic evolution of historical state tokens \( x_{i} \), where \( x \) represents either channel or spatiotemporal states.
    }
    \label{fig_Mamba}
\end{figure}

To model the historical state sequence more robustly, we choose to implement the state reasoning process based on Mamba \cite{mamba}. As an advanced state space model \cite{ssm1,ssm2}, Mamba can effectively capture long-range dependencies in sequential data. Its core selective mechanism dynamically adjusts processing strategies based on input variations, giving the model context awareness and adaptive weight adjustment capabilities, which enhance its robustness in modeling continuous target state transitions.
As shown in Fig.\ref{fig_Mamba},  we append the initial tokens \(\{\mathcal{S}^{init},\mathcal{C}^{init}\}\) to the end of the historical state sequence \(\Psi\) to initialize the state tokens for the current frame. Mamba model then models the temporal variations in the historical state sequence \(\Psi\) and infers the target feature states \(\{\widehat{\mathcal{S}}^{T}, \widehat{\mathcal{C}}^{T}\}\) for the current frame. These states are subsequently used to guide the feature reconstruction process.
This state reasoning process can be expressed as:
\begin{equation}
\begin{split}
&\{\widehat{\mathcal{S}}^{T}, \widehat{\mathcal{C}}^{T}\} \leftarrow 
\xi([\Psi_s \  \Vert \ \mathcal{S}^{init}],[\Psi_c \  \Vert \  \mathcal{C}^{init}] )\\
\end{split}
\end{equation}
\(\xi\) consists of two components: the \(\xi_c\) Mamba network for channel states and the \(\xi_s\) Mamba network for spatial states. Both networks share the same layer types, with channel dimensions adapted to the input. The notation \([a \Vert b]\) denotes concatenation along the sequence dimension. \(\Psi_c\) and \(\Psi_s\) represent the channel and spatial state tokens in the historical state sequence \(\Psi\), respectively.

\subsubsection{\textbf{Temporal Decoder.}} 
The temporal decoder employs a dual-stage cross-attention mechanism to enhance target representations. As shown in Fig.\ref{fig_Overall_framework}(a), the module first concatenates the predicted feature \(\widehat{\mathcal{F}}_{tg}^t\) with the template feature \(\mathcal{F}_z\) along the sequence dimension, constructing a joint reference for target representation. In the first stage of cross-attention, the search region feature \(\mathcal{F}_x\) is used as the query and fused with the template and predicted features, generating a joint query that combines initial, predicted, and current information. 
Then, in the second stage, the joint query further enhances the target representation within the search region. 
The resulting representation is then processed through a feedforward neural network to produce the final output.
Through this dual-stage attention strategy, the temporal decoder combines historical and current information to deeply optimize the target representation.

\subsection{Reasoning Process Supervision.}
To effectively learn contextual reasoning processes and leverage the consistency of targets in video sequences to infer the current target state, we design a forward supervision signal to constrain this process. As show in Fig.\ref{fig_Overall_framework}(b), we first crop the search region target to match the size of the template image using ground truth labels, and then input it along with the template image into the backbone network to extract features, obtaining the true target feature \(\mathcal{F}_{tg}^t\). 
To avoid affecting the optimization process of the backbone network when obtaining the ground truth labels, we choose to freeze this part.
Subsequently, \(\mathcal{F}_{tg}^t\) is passed through a compression module to generate the current true state tokens \(\{\mathcal{S}^{T}, \mathcal{C}^{T}\}\).
To constrain the reasoning process of the state reasoning model, we compute the L2 loss between the true state tokens \(\{\mathcal{S}^{T}, \mathcal{C}^{T}\}\) and the predicted state tokens \(\{\widehat{\mathcal{S}}^{T}, \widehat{\mathcal{C}}^{T}\}\). This loss is expressed as:
\begin{equation}
\mathcal{L}_{\text{state}} = \left\| \mathcal{S}^{T} - \widehat{\mathcal{S}}^{T} \right\|_2^2 + \left\| \mathcal{C}^{T} - \widehat{\mathcal{C}}^{T} \right\|_2^2
\end{equation}
where \(\left\| \cdot \right\|_2\) denotes the L2 norm.
However, the true state tokens \(\{\mathcal{S}^{T}, \mathcal{C}^{T}\}\) are generated through a compression module, and their validity depends on whether the compressed state can accurately reconstruct the original features. 
Therefore, we introduce an additional reconstruction loss to ensure effectiveness between compression and reconstruction. Specifically, we compute the L2 loss between the true target feature \(\mathcal{F}_{tg}^t\) and the reconstructed feature \(\widehat{\mathcal{F}}_{tg}^t\). This loss is expressed as:
\begin{equation}
\mathcal{L}_{\text{recon}} = \left\| \mathcal{F}_{tg}^t - \widehat{\mathcal{F}}_{tg}^t \right\|_2^2
\end{equation}
Unlike a simple compression-reconstruction loss, this loss serves a dual purpose:  
1)Constraining the contextual reasoning process: When historical state sequences \(\Psi\) and initial target features \(\mathcal{F}_z\) are input, the model can infer the current target features.  
2)Supervising the reconstruction process: When the inferred \(\{\widehat{\mathcal{S}}^{T}, \widehat{\mathcal{C}}^{T}\}\) is accurate, the reconstruction network can recover the original target feature \(\mathcal{F}_{tg}^t\) from the compressed state.  
Through this design, we ensure that both the reasoning and reconstruction stages are effectively constrained. The final temporal supervision loss is defined as:
\begin{equation}
\mathcal{L}_{\text{ssm}} = \alpha \mathcal{L}_{\text{state}} + \beta \mathcal{L}_{\text{recon}}
\end{equation}
where \(\alpha = 0.5\) and \(\beta = 1\).

\subsection{ Training and Inference}
\label{subsec: prediction head and loss function}
\textbf{Training.}
We use a mainstream prediction head, which consists of a classification branch and two regression branches. The classification branch predicts the probability of the target's center point at each pixel location and is optimized using focal loss \cite{focal_loss}. The two regression branches predict the width and height of the target, optimized through L1 loss. 
Finally, by combining all predictions, the final predicted bounding box is formed and constrained using GIoU loss \cite{giou_loss}.
The overall loss function is defined as follows:
\begin{equation}
\mathcal{L}_{\text{total}} = \mathcal{L}_{\text{cls}} + \lambda_{iou}\mathcal{L}_{\text{iou}} + \lambda_{l_1}\mathcal{L}_{l_1} + \lambda_{ssm}\mathcal{L}_{\text{ssm}}
\end{equation}
Here, \(\lambda_{iou} = 2\) and \(\lambda_{l_1} = 5\) are regularization parameters consistent with OSTrack \cite{ostrack}, and \(\lambda_{ssm} = 4\).
During training, we maintain the historical state sequence without relying on the predictions from the state inference model, using only feature compression tokens. This approach prevents the accumulation and propagation of prediction errors and allows the model to focus more on learning the reasoning mechanism of the current state.

\textbf{Inference.}
During testing, the historical sequence is maintained using state inference and feature compression. Relying solely on state inference can introduce bias, while feature compression increases computational cost. 
To reduce both, we alternate between these methods. 
Additionally, due to potential tracking failures or reasoning biases, the historical state sequence may contain noise. 
To mitigate this, we use the maximum classification score in the head network as the confidence score. When the score is low, we replace the current state with the initial value to avoid erroneous information. We also model the target reasoning process in reverse from the current to the initial moment within a specified time interval. By summing the forward and backward inferred states, we compensate for missing intermediate states. The score threshold and frame interval are set to 0.4 and 60.

\begin{table}[t]
\centering
\caption{Comparison of model parameters, FLOPs, and Speed.}
\begin{adjustbox}{valign=c,max width=\linewidth}
\begin{tabular}{lcccccc}
\toprule
\textbf{Tracker} & \textbf{Resolution} & \textbf{Params} & \textbf{FLOPs} & \textbf{Speed} & \textbf{Device} \\
\midrule
SeqTrack  & 384$\times$384  & 89M & 148G & 17.8fps & Tesla V100 \\
Ours       & 384$\times$384  & 76M & 57G & 34.7fps & Tesla V100 \\
\bottomrule
\end{tabular}
\end{adjustbox}
\label{tab_speed}
\end{table} 

\section{Experiments}
In this section, we first introduce the experimental details and results on multiple benchmarks, comparing them with SOTA methods. Then, we validate the effectiveness of the modules and strategies through ablation studies. Finally, we visually analyze tracking results and attention maps to understand how RSTrack works.

\subsection{Implementation Details}
\label{subsec: Implementation Details}
The tracker is implemented using Python 3.9 and PyTorch 1.13.1, with training and testing performed on two NVIDIA A100. 
The tracker is evaluated on two NVIDIA Tesla V100 during the tracking speed test phase. The results are presented in Tab.\ref{tab_speed}.

\textbf{Training strategy.}
We train RSTrack on four mainstream datasets: GOT-10k \cite{got10k}, LaSOT \cite{LaSOT}, TrackingNet \cite{TrackingNet}, and COCO \cite{coco}. To ensure effective state reasoning and fair training, we sample 4 search images per video and 15,000 sequences per epoch, totaling 60,000 search images, consistent with mainstream sampling strategies \cite{ostrack, aqatrack}.
For the network architecture, we adopt Fast iTPN-B224 \cite{Fast-iTPN} as the backbone network. In the state reasoning module, the $\xi_c$ and $\xi_s$ Mamba networks and the temporal decoder are all set to 3 iterative layers. 
Regarding training configuration, we use the AdamW optimizer \cite{AdamW} with initial learning rates of \(4\times10^{-5}\) for the backbone network and \(4\times10^{-4}\) for other modules. The model is trained for 150 epochs, with learning rate decay starting at epoch 120 (decay factor \(10^{-4}\)). When training solely on the GOT-10k, we train for 100 epochs, decaying from epoch 80.
For RSTrack-256, the batch size is set to 8, meaning each GPU processes 8 video sequences per iteration, yielding 64 search frames across 2 GPUs. Due to GPU memory limits, RSTrack-384 adopts a batch size of 6, sampling 48 search frames per iteration.

\begin{figure}[t]
    \centering
        \includegraphics[width=\linewidth]{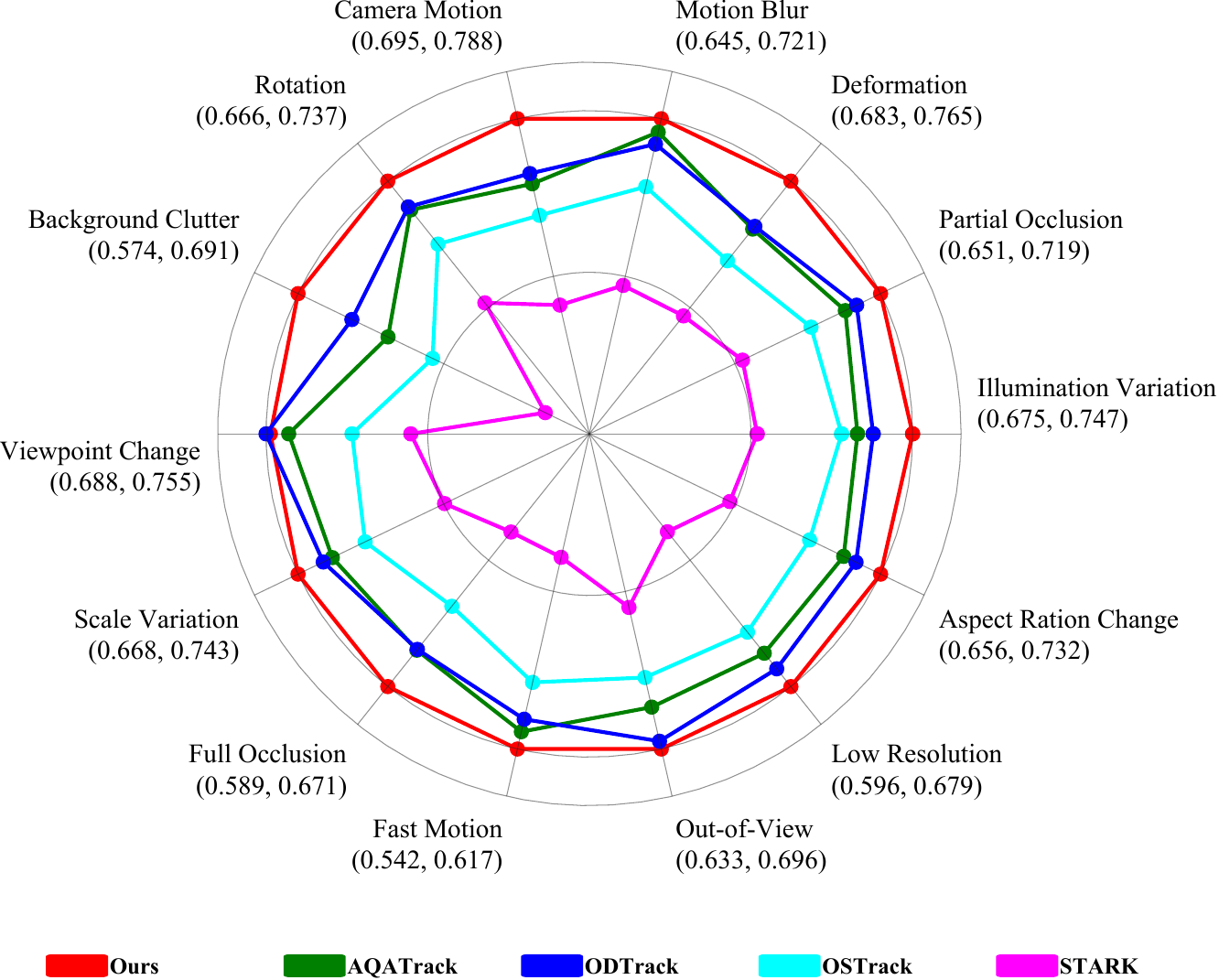}
    \caption{
    \textbf{AUC scores of different attributes on LaSOT.}
    }
    \label{fig_attributes_auc}
\end{figure}
\begin{table*}[t]
    \centering
    \caption{ 
    Comparison with state-of-the-art trackers on four popular benchmarks: LaSOT, LaSOT\textsubscript{ext}, GOT-10K, and TrackingNet. Where $\ast$ denotes for trackers only trained on GOT-10K. Best is shown in \textbf{bold}, second best in \underline{underlined}.
    }
    
    \begin{adjustbox}{valign=c,max width=\textwidth}
        \fontsize{10}{11}\selectfont
        \begin{tabular}{r|c|ccc|ccc|ccc|ccc}
        \toprule
        \multicolumn{1}{c|}{\multirow{2}{*}{Method} }
        & \multicolumn{1}{c|}{\multirow{2}{*}{Source}} 
        & \multicolumn{3}{c|}{LaSOT} 
        & \multicolumn{3}{c|}{LaSOT$_{ext}$} 
        & \multicolumn{3}{c|}{GOT-10k$^*$} 
        & \multicolumn{3}{c}{TrackingNet}\\
        \cline{3-14}
        && AUC & P$_{norm}$ & P     & AUC & P$_{norm}$ & P      & AO & SR$_{0.5}$ & SR$_{0.75}$     & AUC & P$_{norm}$ & P \\
        \midrule
        \textbf{{RSTrack}-256} & Ours & 
        \textbf{73.4} & \textbf{84.1} & \textbf{81.7} &
        \textbf{52.7} & \underline{63.8} & \textbf{60.3} & 
        \textbf{76.6} & \textbf{86.4} & \textbf{76.9} & 
        \textbf{85.3} & \textbf{89.9} & \textbf{84.7}\\
        \midrule
        TemTrack-256\cite{temtrack} & AAAI25   & \underline{72.0} & 82.1 & 79.1     & \underline{52.4} & 63.3 & \underline{60.2} 
        & 74.9 & 84.8 & 71.7        & 84.3 & 88.8 & 83.5\\

        MambaLCT-256\cite{mambalct} & AAAI25  & 71.8 & \underline{83.0} & \underline{79.4}    & 51.6 & \textbf{64.0} & 59.0  & 74.8 & \underline{85.4} & 72.1    & 84.3 & 89.2 & 83.9\\

        ARTrackV2-256\cite{artrackv2} &CVPR24   & 71.6 & 80.2 & 77.2    & 50.8 & 61.9 & 57.7    
        & \underline{75.9} & \underline{85.4} & \underline{72.7}    & \underline{84.9} & \underline{89.3} & \underline{84.5} \\

        AQATrack-256\cite{aqatrack} & CVPR24    & 71.4 & 81.9 & 78.6    & 51.2 & 62.2 & 58.9    & 73.8 & 83.2 & 72.1    
        & 83.8 & 88.6 & 83.1\\

        EVPTrack-224\cite{evptrack} &AAAI24     & 70.4 & 80.9 & 77.2    & 48.7 & 59.5 & 55.1    & 73.3 & 83.6 & 70.7    
        & 83.5 & 88.3 & - \\
        F-BDMTrack-256\cite{F-BDMTrack}&ICCV23  & 69.9 & 79.4 & 75.8    & 47.9 & 57.9 & 54.0    & 72.7 & 82.0 & 69.9    & 83.7 & 88.3 & 82.6 \\
        SeqTrack-B256\cite{seqtrack} & CVPR23   & 69.9 & 79.7 & 76.3    & 49.5 & 60.8 & 56.3    & 74.7 & 84.7 & 71.8    
        & 83.3 & 88.3    & 82.2 \\
        MixFormer-22k\cite{mixformer}&CVPR22    & 69.2 & 78.7 & 74.7    & -    & -    & -       & 70.7 & 80.0 & 67.8    
        & 83.1 & 88.1 & 81.6 \\
        OSTrack-256\cite{ostrack} &ECCV22       & 69.1 & 78.7 & 75.2    & 47.4 & 57.3 & 53.3    & 71.0 & 80.4 & 68.2   
        & 83.1 & 87.8 & 82.0\\
        STARK-ST101\cite{stark} & ICCV21        & 67.1 & 77.0 & -       & -    &-     & -       & 68.8 & 78.1 & 64.1    
        & 82.0 & 86.9 & -    \\
        TransT \cite{transt}& CVPR21            & 64.9 & 73.8 & 69.0    & -    & -    & -       & 67.1 & 76.8 & 60.9    
        & 81.4 & 86.7 & 80.3 \\
        Ocean \cite{Ocean}&  ECCV 20            & 56.0 & 65.1 & 56.6    &-     & -    &-        & 61.1 & 72.1 & 47.3    
        & -    & -    &-     \\
        SiamRPN++\cite{siamrpn++}&CVPR19        & 49.6 & 56.9 & 49.1    & 34.0 & 41.6 & 39.6    & 51.7 & 61.6 & 32.5    
        & 73.3 & 80.0 & 69.4 \\
        ECO \cite{ECO} & ICCV 17                & 32.4 & 33.8 & 30.1    & 22.0 & 25.2 & 24.0    & 31.6 & 30.9 & 11.1    
        & -    & -    & - \\
        SiamFC \cite{siamfc} & ECCVW16          & 33.6 & 42.0 & 33.9    & 23.0 & 31.1 & 26.9    & 34.8 & 35.3 & 9.8     
        & -    & -    & - \\
        \midrule
        
        \multicolumn{14}{l}{\multirow{1}{*}{\textit{Some Trackers with Higher Resolution}} }\\
        \midrule
        OSTrack-384\cite{ostrack}&ECCV22        & 71.1 & 81.1 & 77.6    & 50.5 & 61.3 & 57.6    & 73.7 & 83.2 & 70.8    & 83.9 & 88.5 & 83.2  \\
        ROMTrack-384\cite{ROMTrack} & ICCV23    & 71.4 & 81.4 & 78.2    & 51.3 & 62.4 & 58.6    & 74.2 & 84.3 & 72.4    & 84.1 & 89.0 & 83.7 \\
        SeqTrack-B384\cite{seqtrack} & CVPR23   & 71.5 & 81.1 & 77.8    & 50.5 & 61.6 & 57.5    & 74.5 & 84.3 & 71.4    & 83.9 & 88.8 & 83.6 \\
        AQATrack-384\cite{aqatrack} & CVPR24    & 72.7 & 82.9 & 80.2    & 52.7 & 64.2 & 60.8    & 76.0 & 85.2 & 74.9  & 84.8 & 89.3 & 84.3\\
        ARTrackV2-B384\cite{artrackv2} & CVPR24     & 73.0 & 82.0 & 79.6    & 52.9 & 63.4 & 59.1    & \underline{77.5} & 86.0 & \underline{75.5}    & \underline{85.7} & \underline{89.8} & \underline{85.5}\\
        TemTrack-384\cite{temtrack} & AAAI25    & 73.1 & 83.0 & 80.7   & \underline{53.4}  & \underline{64.8} & 61.0    
        & 76.1 & 84.9 & 74.4   & 85.0 & 89.3 & 84.8\\
        MambaLCT-384\cite{mambalct} & AAAI25    & \underline{73.6} & \underline{84.1} & \underline{81.6}  & 53.3 & \underline{64.8} & \underline{61.4}    & 76.2 & \underline{86.7} & 74.3  & 85.2 & \underline{89.8} & 85.2\\
        \midrule
        \textbf{{RSTrack}-384} 
        & Ours & \textbf{74.4} & \textbf{84.2} & \textbf{83.0}    
        & \textbf{53.9} & \textbf{65.5} & \textbf{61.8}    
        & \textbf{78.1} & \textbf{87.1} & \textbf{78.9} 
        & \textbf{85.8} & \textbf{90.3} & \textbf{85.7}\\
        \bottomrule    
        \end{tabular}
        \end{adjustbox}

    \label{tab_performance_1}
\end{table*} 

\begin{table*}[t]
\centering
\caption{Comparison with state-of-the-art trackers on TNL2K and UAV123 benchmarks in AUC score.}
\begin{adjustbox}{valign=c,max width=\textwidth}
\begin{tabular}{l|cccccccccc|c}
\toprule
\textbf{Tracker} & \textbf{SiamFC} & \textbf{ECO} & \textbf{SiamRPN++} & \textbf{TransT} & \textbf{OSTrack} & \textbf{SeqTrack} & \textbf{F-BDMTrack} & \textbf{EVPTrack} & \textbf{ARTrackV2} & \textbf{MambaLCT} & \textbf{RSTrack} \\
\midrule
\textbf{UAV123} & 46.8 & 53.5 & 61.0 & 69.1 & 68.3 & 69.2 & 69.0 & 70.2 & 69.9 & 70.1 & \textbf{70.5} \\
\textbf{TNL2K} & 29.5 & 32.6 & 41.3 & 50.7 & 54.3 & 54.9 & 56.4 & 57.5 & 59.2 &58.5 & \textbf{60.5} \\
\bottomrule
\end{tabular}
\end{adjustbox}
\label{tab_performance_2}
\end{table*}

\subsection{Comparison with State-of-the-arts}
\label{subsec: Comparison with State-of-the-arts}
We compare our evaluation results with other SOTA methods on six benchmarks to prove our effectiveness.

\textbf{LaSOT.}
LaSOT \cite{LaSOT} is a large-scale long-term tracking benchmark with 280 test videos averaging 2,448 frames. As shown in Tab.\ref{tab_performance_1}, RSTrack outperforms fifteen mainstream trackers across all metrics. Benefiting from the context reasoning supervision mechanism, RSTrack precisely captures the dynamic evolution of target appearance, exhibiting strong performance in challenging scenarios like fast motion, deformation, and full occlusion (see Fig.\ref{fig_attributes_auc}).

\textbf{\(\text{LaSOT}_{ext}\).}
\(\text{LaSOT}_{ext}\) \cite{lasot_ext}, an extension of LaSOT \cite{LaSOT}, introduces more challenging scenarios, such as fast-moving small objects. As shown in Tab.\ref{tab_performance_1}, compared to TemTrack-384 with the same backbone and input resolution, RSTrack has achieved a performance improvement of at least 0.5\% across all three evaluation metrics, further substantiating the superiority of our proposed method.

\textbf{GOT-10k.}
GOT-10k \cite{got10k} test set consists of 180 videos covering common tracking challenges.
Following the official protocol, we train our models using only the GOT-10k training set.
As shown in Tab.\ref{tab_performance_1}, our method outperforms all SOTA trackers across all metrics.
Thanks to the effective supervision of target state reasoning, RSTrack-256 achieves 76.9\% on SR$_{0.75}$, surpassing ARTrackV2-256 by 4.2\%, demonstrating strong capability in modeling target states.

\textbf{TrackingNet.}
TrackingNet \cite{TrackingNet} is a large-scale short-term tracking benchmark, comprising over 30,000 training sequences and 511 testing sequences, covering a wide range of real-world scenarios and content, as shown in Tab.\ref{tab_performance_1}. We present the results of RSTrack and some SOTA trackers on TrackingNet. RSTrack-256 achieves the best AUC score of 85.3\%, surpassing TemTrack-256 by 1\%.

\textbf{TNL2K and UAV123.} 
We further evaluate RSTrack on two tracking benchmarks: TNL2K \cite{tnl2k}, a large-scale dataset for natural language tracking consisting of 700 test videos, and UAV123 \cite{uav123}, a low-altitude drone tracking dataset with 123 videos. As shown in Tab.\ref{tab_performance_2}, RSTrack-256 achieves 70.5\% AUC on UAV123 and 60.5\% on TNL2K, outperforming all existing trackers.

\subsection{Ablation and Analysis}
\label{subsec: Ablation Study}
In this section, we use RSTrack-B256 as the baseline model in our ablation study. The result of the baseline is reported in Tab.\ref{tab_ablation_var}.

\subsubsection{\textbf{Network Architectures Variants}}
In this section, we compare the implementation methods' impact on state evolution.

\textbf{\textit{Compress v.s. Aggregate.}}
We compared the method of modeling temporal evolution by transmitting contextual tokens, which is similar to TemTrack \cite{temtrack}. The results (Tab.\ref{tab_ablation_var}(\#2)) show that the transfer token method performs 1.2\% worse than the compressed state token method. This difference might be because compressed tokens can learn the target's core information more effectively, while transfer tokens, receiving too much diverse information, may inadequately capture the current frame's unique feature.

\textbf{\textit{Mamba v.s. ViT.}}
We tried to replace Mamba \cite{mamba} with ViT \cite{vit}, applying an upper triangular mask to enforce temporal order (i.e., earlier states cannot access later ones). The results (Tab.\ref{tab_ablation_var} (\#3)) show that Mamba \cite{mamba} performs better, indicating that its data dependency characteristics better handle dynamic contextual changes.

\textbf{\textit{SCM v.s. CBAM.}} 
Our compression module is designed to extract key target information from both spatial and channel dimensions. To evaluate its effectiveness, we conducted comparative experiments with the spatial and channel attention mechanisms in CBAM \cite{ca_sa}. The results (Tab.\ref{tab_ablation_var} (\#4)) show that our compression module achieves superior performance.

\subsubsection{\textbf{Component Ablation Analysis}}
We evaluate the impact of modules or loss functions on tracker performance by selectively removing them.  
Tab.\ref{tab_ablation_rm} (\#1) shows the baseline performance using only the visual encoder and prediction head.  
The results (Tab.\ref{tab_ablation_rm} (\#2)) show that removing all context reasoning components (only keeping the decoder) reduces tracking success by 1.6\%, indicating that the predicted target features from state reasoning modeling effectively enhance search region features.
Further, removing the decoder and using cross-correlation \cite{siamfc} instead lowers success by 1.4\%.  
Notably, without SSM loss for temporal modeling (Tab.\ref{tab_ablation_rm} (\#4)), success drops by 1\%.  
This suggests that explicitly modeling and supervising the target's temporal association achieves better performance than implicitly stacking historical information.

\begin{figure}[t]
    \centering
        \includegraphics[width=\linewidth]{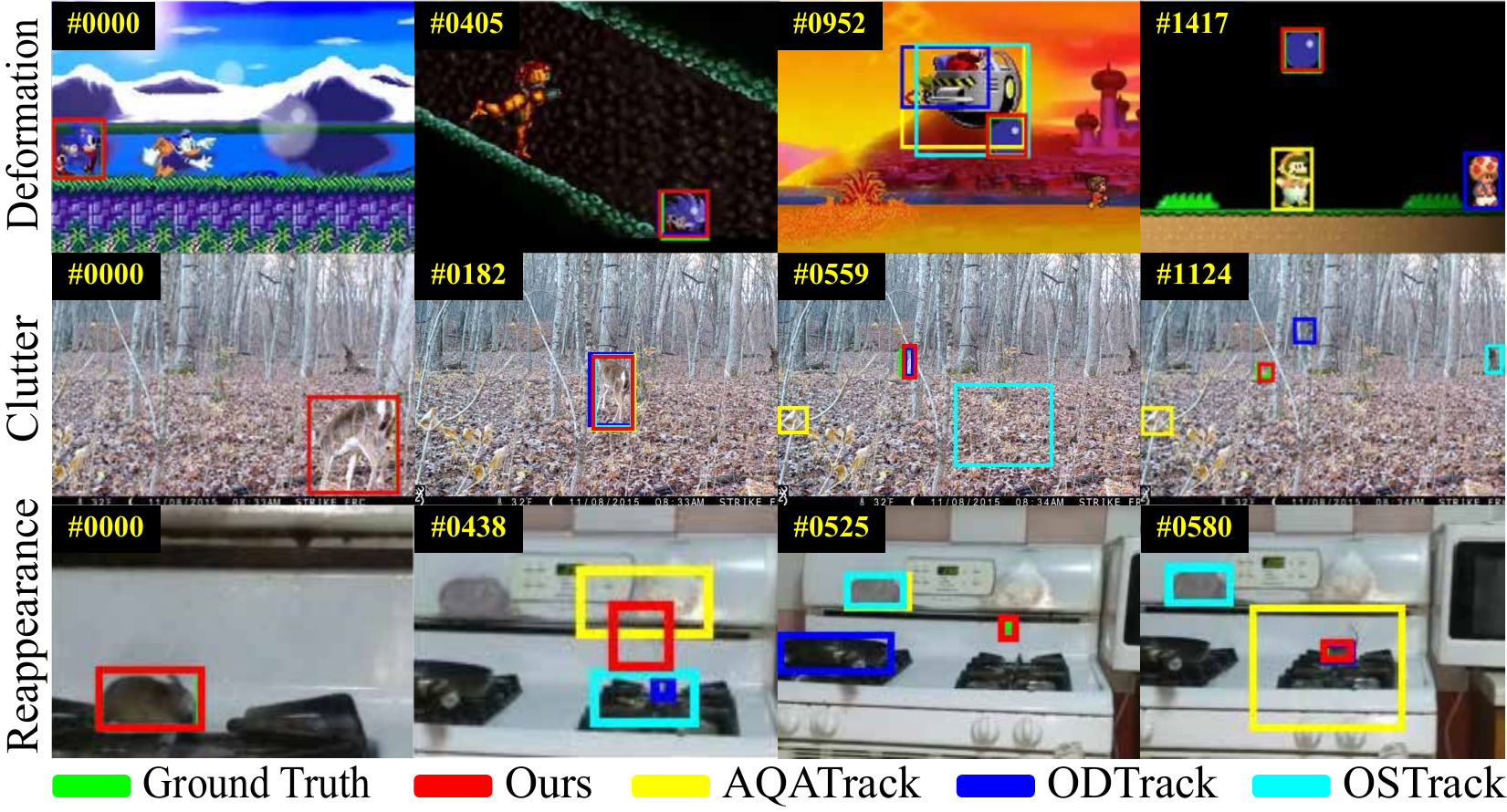}
    \caption{
    \textbf{Visual Comparison. Our tracker \textit{vs.} three SOTA trackers on LaSOT under three challenges.}
    }
    \label{fig_visualization_bbox}
\end{figure}
\begin{table}[t]
\centering
\caption{
Ablation study on LaSOT. We use \textcolor{gray}{gray},
\textcolor{green}{green}, and \textcolor{yellow}{yellow} to denote RSTrack, network architectures variants, and history state integration, respectively. $\Delta$  denotes the performance change in AUC compared to RSTrack.
}

\resizebox{\linewidth}{!}{ 
\begin{tabular}{c l c c c c}
\toprule
\# & \multicolumn{1}{|l|}{Method} & AUC & P$_{norm}$ &\multicolumn{1}{c|}{P} & $\Delta$ \\ 
\midrule
\rowcolor{gray!15}
\multicolumn{1}{l|}{1} & RSTrack 
& \multicolumn{1}{|c}{73.4} & 84.1 & 81.7 & \multicolumn{1}{|c}{-} \\ 

\rowcolor{green!5} 
\multicolumn{1}{l|}{2} & Compress $\rightarrow$ Aggregate
& \multicolumn{1}{|c}{72.2} & 82.4 & 79.6 & \multicolumn{1}{|c}{-1.2}  \\ 

\rowcolor{green!5}
\multicolumn{1}{l|}{3} & Mamba $\rightarrow$ ViT 
& \multicolumn{1}{|c}{71.9} & 82.2 & 79.3 & \multicolumn{1}{|c}{-1.5} \\ 

\rowcolor{green!5} 
\multicolumn{1}{l|}{4} &  SCM $\rightarrow$ CBAM
&  \multicolumn{1}{|c}{72.8} & 83.1 & 80.6 & \multicolumn{1}{|c}{-0.6} \\ 

\rowcolor{yellow!5}
\multicolumn{1}{l|}{5} & Windows Sample $\rightarrow$ Global Sample
&  \multicolumn{1}{|c}{72.9} & 83.5 & 81.2 & \multicolumn{1}{|c}{-0.5}  \\ 

\rowcolor{yellow!5} 
\multicolumn{1}{l|}{6} & Bidirectional  $\rightarrow$  Unidirectional
&  \multicolumn{1}{|c}{73.0} & 83.5 & 81.0 & \multicolumn{1}{|c}{-0.4} \\ 
\bottomrule

\end{tabular}
}
\label{tab_ablation_var}
\end{table}

\subsubsection{\textbf{History State Integration}}
In this section, we explore the sampling strategy of target historical states and additional reasoning directions, analyzing their impact on target state modeling.

\textbf{\textit{Global Sample v.s. Windows Sample.}}
We compared two sampling strategies for historical state sequences: sampling the entire history (Global Sample) and window sampling the most recent state tokens (Windows Sample).
Experimental results show that window sampling improves performance by about 0.5\% compared to global sampling (Tab.\ref{tab_ablation_var} (\#5)). 
This may be due to window sampling’s better capture of the target’s latest changes, while global sampling processes all video frame states, making it harder to model current target changes accurately. The window size is set to 500.

\textbf{\textit{Bidirectional v.s. Unidirectional.}}
We reverse the historical state sequence, achieving bidirectional modeling and verification of target behavior. Experimental results (Tab.\ref{tab_ablation_var} (\#6)) show that this bidirectional dynamic reasoning method outperforms the unidirectional approach, capturing the temporal features and dynamic patterns of target behavior more comprehensively.

\subsubsection{\textbf{Visualization}}
\begin{figure}[t]
    \centering
        \includegraphics[width=\linewidth]{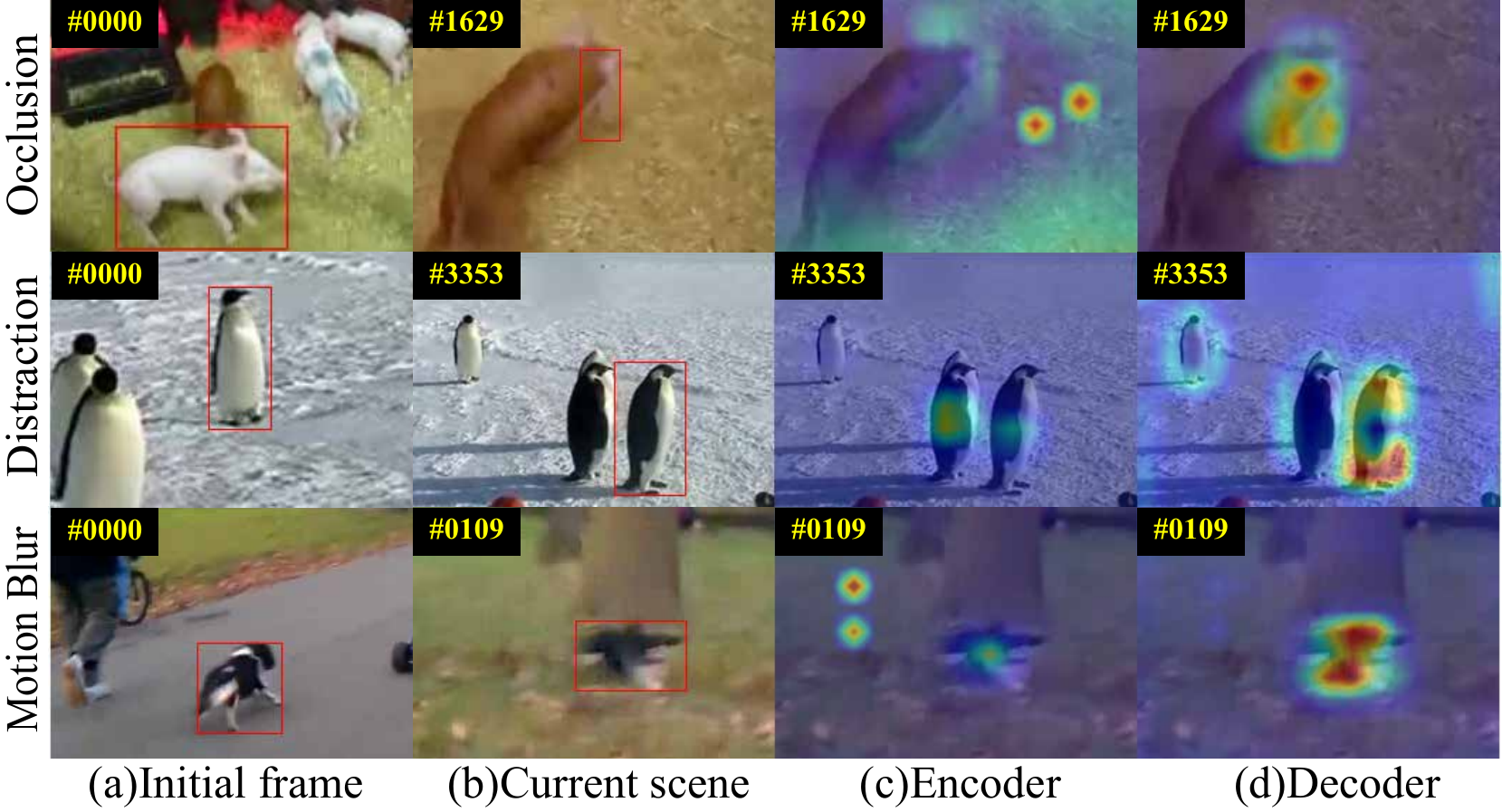}
    \caption{
    Attention Visualization. (a): Initial template frame. (b): current search region. The \textcolor{red}{red} boxes denote the ground truth. (c): Cross-attention map in the visual encoder. (d): Cross-attention map in the temporal decoder.
    }
    \label{fig_visualization_attn}
\end{figure}

We compared RSTrack with three state-of-the-art trackers: OSTrack \cite{ostrack}, ODTrack \cite{odtrack}, and AQATrack \cite{aqatrack} across three challenging scenarios.
As shown in Fig.\ref{fig_visualization_bbox}, the experimental results demonstrate that RSTrack exhibits outstanding adaptability in complex environments where reliance on static appearance is difficult.
Additionally, we visualized the attention maps of the visual encoder and temporal decoder on the search frames (see Fig.\ref{fig_visualization_attn}) and conducted a comparative analysis across the three challenging scenarios. The results indicate that when static template frames struggle to accurately distinguish the target in complex scenarios, the temporal features inferred through contextual reasoning can more effectively capture the target information, demonstrating superior robustness.

\begin{table}[t]
\centering
\caption{
Ablation study for important components. Blank denotes the component used by default, while \ding{55} represents the component removed. Performance is evaluated on LaSOT.
}
\resizebox{\linewidth}{!}{ 
\begin{tabular}{c|c c c c |c c c}
\toprule
\# & SRM & TD & RE & \(\mathcal{L}_{\text{ssm}}\) & AUC(\%) & P$_{norm}$(\%) & P(\%) \\ 
\midrule
1 & \ding{55} & \ding{55} & \ding{55} & \ding{55} &  71.1 & 81.2 & 78.2 \\ 
2 & \ding{55} &  & \ding{55} & \ding{55} &  71.8 & 82.3 & 79.8 \\ 
3 &   & \ding{55} &   &  &  72.0 & 82.3 & 79.8 \\ 
4 &   &  &   & \ding{55} &  72.4 & 82.6 & 80.2 \\ 
\rowcolor{gray!15}
5 &  &  &  &  &  \textbf{73.4} & \textbf{84.1} & \textbf{81.7} \\ 
\bottomrule
\end{tabular}
}
\label{tab_ablation_rm}
\end{table}
\section{Conclusion}
In this paper, we present RSTrack, an innovative context learning framework. By incorporating supervised signals into the context modeling process, we explicitly learn the temporal variations of the target state over history and reason about the current state based on this, thereby achieving more robust video-level context modeling.Extensive experimental results demonstrate that our method performs excellently across six mainstream tracking benchmarks. We hope that this research provides valuable insights for existing context learning in visual tracking and contributes to the further development of this field.

\begin{acks}
This work is supported by the Project of Guangxi Science and Technology (No.2025GXNSFAA069676 and 2024GXNSFGA010001), the National Natural Science Foundation of China (No.U23A20383, 62472109, 62466051 and 62402252), the Guangxi “Young Bagui Scholar” Teams for Innovation and Research Project, the Research Project of Guangxi Normal University (No.2024DF001), the grant from Guangxi Colleges and Universities Key Laboratory of Intelligent Software (No.2024B01).
\end{acks}
\bibliographystyle{ACM-Reference-Format}
\balance
\bibliography{bib/mine}


\begin{thebibliography}{52}


\ifx \showCODEN    \undefined \def \showCODEN     #1{\unskip}     \fi
\ifx \showISBNx    \undefined \def \showISBNx     #1{\unskip}     \fi
\ifx \showISBNxiii \undefined \def \showISBNxiii  #1{\unskip}     \fi
\ifx \showISSN     \undefined \def \showISSN      #1{\unskip}     \fi
\ifx \showLCCN     \undefined \def \showLCCN      #1{\unskip}     \fi
\ifx \shownote     \undefined \def \shownote      #1{#1}          \fi
\ifx \showarticletitle \undefined \def \showarticletitle #1{#1}   \fi
\ifx \showURL      \undefined \def \showURL       {\relax}        \fi
\providecommand\bibfield[2]{#2}
\providecommand\bibinfo[2]{#2}
\providecommand\natexlab[1]{#1}
\providecommand\showeprint[2][]{arXiv:#2}

\bibitem[Bai et~al\mbox{.}(2024)]%
        {artrackv2}
\bibfield{author}{\bibinfo{person}{Yifan Bai}, \bibinfo{person}{Zeyang Zhao}, \bibinfo{person}{Yihong Gong}, {and} \bibinfo{person}{Xing Wei}.} \bibinfo{year}{2024}\natexlab{}.
\newblock \showarticletitle{Artrackv2: Prompting autoregressive tracker where to look and how to describe}. In \bibinfo{booktitle}{\emph{Proceedings of the IEEE/CVF conference on computer vision and pattern recognition}}. \bibinfo{pages}{19048--19057}.
\newblock


\bibitem[Benchmark(2016)]%
        {uav123}
\bibfield{author}{\bibinfo{person}{UT Benchmark}.} \bibinfo{year}{2016}\natexlab{}.
\newblock \showarticletitle{A benchmark and simulator for uav tracking}. In \bibinfo{booktitle}{\emph{European conference on computer vision}}, Vol.~\bibinfo{volume}{7}.
\newblock


\bibitem[Bertinetto et~al\mbox{.}(2016)]%
        {siamfc}
\bibfield{author}{\bibinfo{person}{Luca Bertinetto}, \bibinfo{person}{Jack Valmadre}, \bibinfo{person}{Joao~F Henriques}, \bibinfo{person}{Andrea Vedaldi}, {and} \bibinfo{person}{Philip~HS Torr}.} \bibinfo{year}{2016}\natexlab{}.
\newblock \showarticletitle{Fully-convolutional siamese networks for object tracking}. In \bibinfo{booktitle}{\emph{Computer Vision--ECCV 2016 Workshops: Amsterdam, The Netherlands, October 8-10 and 15-16, 2016, Proceedings, Part II 14}}. Springer, \bibinfo{pages}{850--865}.
\newblock


\bibitem[Cai et~al\mbox{.}(2023)]%
        {ROMTrack}
\bibfield{author}{\bibinfo{person}{Yidong Cai}, \bibinfo{person}{Jie Liu}, \bibinfo{person}{Jie Tang}, {and} \bibinfo{person}{Gangshan Wu}.} \bibinfo{year}{2023}\natexlab{}.
\newblock \showarticletitle{Robust object modeling for visual tracking}. In \bibinfo{booktitle}{\emph{Proceedings of the IEEE/CVF international conference on computer vision}}. \bibinfo{pages}{9589--9600}.
\newblock


\bibitem[Chen et~al\mbox{.}(2023)]%
        {seqtrack}
\bibfield{author}{\bibinfo{person}{Xin Chen}, \bibinfo{person}{Houwen Peng}, \bibinfo{person}{Dong Wang}, \bibinfo{person}{Huchuan Lu}, {and} \bibinfo{person}{Han Hu}.} \bibinfo{year}{2023}\natexlab{}.
\newblock \showarticletitle{Seqtrack: Sequence to sequence learning for visual object tracking}. In \bibinfo{booktitle}{\emph{Proceedings of the IEEE/CVF conference on computer vision and pattern recognition}}. \bibinfo{pages}{14572--14581}.
\newblock


\bibitem[Chen et~al\mbox{.}(2021)]%
        {transt}
\bibfield{author}{\bibinfo{person}{Xin Chen}, \bibinfo{person}{Bin Yan}, \bibinfo{person}{Jiawen Zhu}, \bibinfo{person}{Dong Wang}, \bibinfo{person}{Xiaoyun Yang}, {and} \bibinfo{person}{Huchuan Lu}.} \bibinfo{year}{2021}\natexlab{}.
\newblock \showarticletitle{Transformer tracking}. In \bibinfo{booktitle}{\emph{Proceedings of the IEEE/CVF conference on computer vision and pattern recognition}}. \bibinfo{pages}{8126--8135}.
\newblock


\bibitem[Chen et~al\mbox{.}(2022)]%
        {siamban}
\bibfield{author}{\bibinfo{person}{Zedu Chen}, \bibinfo{person}{Bineng Zhong}, \bibinfo{person}{Guorong Li}, \bibinfo{person}{Shengping Zhang}, \bibinfo{person}{Rongrong Ji}, \bibinfo{person}{Zhenjun Tang}, {and} \bibinfo{person}{Xianxian Li}.} \bibinfo{year}{2022}\natexlab{}.
\newblock \showarticletitle{SiamBAN: Target-aware tracking with Siamese box adaptive network}.
\newblock \bibinfo{journal}{\emph{IEEE Transactions on Pattern Analysis and Machine Intelligence}} \bibinfo{volume}{45}, \bibinfo{number}{4} (\bibinfo{year}{2022}), \bibinfo{pages}{5158--5173}.
\newblock


\bibitem[Cui et~al\mbox{.}(2022)]%
        {mixformer}
\bibfield{author}{\bibinfo{person}{Yutao Cui}, \bibinfo{person}{Cheng Jiang}, \bibinfo{person}{Limin Wang}, {and} \bibinfo{person}{Gangshan Wu}.} \bibinfo{year}{2022}\natexlab{}.
\newblock \showarticletitle{Mixformer: End-to-end tracking with iterative mixed attention}. In \bibinfo{booktitle}{\emph{Proceedings of the IEEE/CVF conference on computer vision and pattern recognition}}. \bibinfo{pages}{13608--13618}.
\newblock


\bibitem[Danelljan et~al\mbox{.}(2017)]%
        {ECO}
\bibfield{author}{\bibinfo{person}{Martin Danelljan}, \bibinfo{person}{Goutam Bhat}, \bibinfo{person}{Fahad Shahbaz~Khan}, {and} \bibinfo{person}{Michael Felsberg}.} \bibinfo{year}{2017}\natexlab{}.
\newblock \showarticletitle{Eco: Efficient convolution operators for tracking}. In \bibinfo{booktitle}{\emph{Proceedings of the IEEE conference on computer vision and pattern recognition}}. \bibinfo{pages}{6638--6646}.
\newblock


\bibitem[Dosovitskiy(2020)]%
        {vit}
\bibfield{author}{\bibinfo{person}{Alexey Dosovitskiy}.} \bibinfo{year}{2020}\natexlab{}.
\newblock \showarticletitle{An image is worth 16x16 words: Transformers for image recognition at scale}.
\newblock \bibinfo{journal}{\emph{arXiv preprint arXiv:2010.11929}} (\bibinfo{year}{2020}).
\newblock


\bibitem[Fan et~al\mbox{.}(2021)]%
        {lasot_ext}
\bibfield{author}{\bibinfo{person}{Heng Fan}, \bibinfo{person}{Hexin Bai}, \bibinfo{person}{Liting Lin}, \bibinfo{person}{Fan Yang}, \bibinfo{person}{Peng Chu}, \bibinfo{person}{Ge Deng}, \bibinfo{person}{Sijia Yu}, \bibinfo{person}{Harshit}, \bibinfo{person}{Mingzhen Huang}, \bibinfo{person}{Juehuan Liu}, {et~al\mbox{.}}} \bibinfo{year}{2021}\natexlab{}.
\newblock \showarticletitle{Lasot: A high-quality large-scale single object tracking benchmark}.
\newblock \bibinfo{journal}{\emph{International Journal of Computer Vision}}  \bibinfo{volume}{129} (\bibinfo{year}{2021}), \bibinfo{pages}{439--461}.
\newblock


\bibitem[Fan et~al\mbox{.}(2019)]%
        {LaSOT}
\bibfield{author}{\bibinfo{person}{Heng Fan}, \bibinfo{person}{Liting Lin}, \bibinfo{person}{Fan Yang}, \bibinfo{person}{Peng Chu}, \bibinfo{person}{Ge Deng}, \bibinfo{person}{Sijia Yu}, \bibinfo{person}{Hexin Bai}, \bibinfo{person}{Yong Xu}, \bibinfo{person}{Chunyuan Liao}, {and} \bibinfo{person}{Haibin Ling}.} \bibinfo{year}{2019}\natexlab{}.
\newblock \showarticletitle{Lasot: A high-quality benchmark for large-scale single object tracking}. In \bibinfo{booktitle}{\emph{Proceedings of the IEEE/CVF conference on computer vision and pattern recognition}}. \bibinfo{pages}{5374--5383}.
\newblock


\bibitem[Fu et~al\mbox{.}(2021)]%
        {stmtrack}
\bibfield{author}{\bibinfo{person}{Zhihong Fu}, \bibinfo{person}{Qingjie Liu}, \bibinfo{person}{Zehua Fu}, {and} \bibinfo{person}{Yunhong Wang}.} \bibinfo{year}{2021}\natexlab{}.
\newblock \showarticletitle{Stmtrack: Template-free visual tracking with space-time memory networks}. In \bibinfo{booktitle}{\emph{Proceedings of the IEEE/CVF conference on computer vision and pattern recognition}}. \bibinfo{pages}{13774--13783}.
\newblock


\bibitem[Gu and Dao(2024)]%
        {mamba}
\bibfield{author}{\bibinfo{person}{Albert Gu} {and} \bibinfo{person}{Tri Dao}.} \bibinfo{year}{2024}\natexlab{}.
\newblock \bibinfo{title}{Mamba: Linear-Time Sequence Modeling with Selective State Spaces}.
\newblock
\showeprint[arxiv]{2312.00752}~[cs.LG]
\urldef\tempurl%
\url{https://arxiv.org/abs/2312.00752}
\showURL{%
\tempurl}


\bibitem[Gu et~al\mbox{.}(2022)]%
        {ssm1}
\bibfield{author}{\bibinfo{person}{Albert Gu}, \bibinfo{person}{Ankit Gupta}, \bibinfo{person}{Karan Goel}, {and} \bibinfo{person}{Christopher Ré}.} \bibinfo{year}{2022}\natexlab{}.
\newblock \bibinfo{title}{On the Parameterization and Initialization of Diagonal State Space Models}.
\newblock
\showeprint[arxiv]{2206.11893}~[cs.LG]
\urldef\tempurl%
\url{https://arxiv.org/abs/2206.11893}
\showURL{%
\tempurl}


\bibitem[Gupta et~al\mbox{.}(2022)]%
        {ssm2}
\bibfield{author}{\bibinfo{person}{Ankit Gupta}, \bibinfo{person}{Albert Gu}, {and} \bibinfo{person}{Jonathan Berant}.} \bibinfo{year}{2022}\natexlab{}.
\newblock \bibinfo{title}{Diagonal State Spaces are as Effective as Structured State Spaces}.
\newblock
\showeprint[arxiv]{2203.14343}~[cs.LG]
\urldef\tempurl%
\url{https://arxiv.org/abs/2203.14343}
\showURL{%
\tempurl}


\bibitem[Hu et~al\mbox{.}(2024a)]%
        {STTrack}
\bibfield{author}{\bibinfo{person}{Xiantao Hu}, \bibinfo{person}{Ying Tai}, \bibinfo{person}{Xu Zhao}, \bibinfo{person}{Chen Zhao}, \bibinfo{person}{Zhenyu Zhang}, \bibinfo{person}{Jun Li}, \bibinfo{person}{Bineng Zhong}, {and} \bibinfo{person}{Jian Yang}.} \bibinfo{year}{2024}\natexlab{a}.
\newblock \showarticletitle{Exploiting Multimodal Spatial-temporal Patterns for Video Object Tracking}.
\newblock \bibinfo{journal}{\emph{arXiv preprint arXiv:2412.15691}} (\bibinfo{year}{2024}).
\newblock


\bibitem[Hu et~al\mbox{.}(2025)]%
        {APTrack}
\bibfield{author}{\bibinfo{person}{Xiantao Hu}, \bibinfo{person}{Bineng Zhong}, \bibinfo{person}{Qihua Liang}, \bibinfo{person}{Zhiyi Mo}, \bibinfo{person}{Liangtao Shi}, \bibinfo{person}{Ying Tai}, {and} \bibinfo{person}{Jian Yang}.} \bibinfo{year}{2025}\natexlab{}.
\newblock \showarticletitle{Adaptive Perception for Unified Visual Multi-modal Object Tracking}.
\newblock \bibinfo{journal}{\emph{arXiv preprint arXiv:2502.06583}} (\bibinfo{year}{2025}).
\newblock


\bibitem[Hu et~al\mbox{.}(2024b)]%
        {MCTrack}
\bibfield{author}{\bibinfo{person}{Xiantao Hu}, \bibinfo{person}{Bineng Zhong}, \bibinfo{person}{Qihua Liang}, \bibinfo{person}{Shengping Zhang}, \bibinfo{person}{Ning Li}, {and} \bibinfo{person}{Xianxian Li}.} \bibinfo{year}{2024}\natexlab{b}.
\newblock \showarticletitle{Toward Modalities Correlation for RGB-T Tracking}.
\newblock \bibinfo{journal}{\emph{IEEE Transactions on Circuits and Systems for Video Technology}} \bibinfo{volume}{34}, \bibinfo{number}{10} (\bibinfo{year}{2024}), \bibinfo{pages}{9102--9111}.
\newblock
\href{https://doi.org/10.1109/TCSVT.2024.3396289}{doi:\nolinkurl{10.1109/TCSVT.2024.3396289}}


\bibitem[Hu et~al\mbox{.}(2024c)]%
        {FFTrack}
\bibfield{author}{\bibinfo{person}{Xiantao Hu}, \bibinfo{person}{Bineng Zhong}, \bibinfo{person}{Qihua Liang}, \bibinfo{person}{Shengping Zhang}, \bibinfo{person}{Ning Li}, \bibinfo{person}{Xianxian Li}, {and} \bibinfo{person}{Rongrong Ji}.} \bibinfo{year}{2024}\natexlab{c}.
\newblock \showarticletitle{Transformer Tracking via Frequency Fusion}.
\newblock \bibinfo{journal}{\emph{IEEE Transactions on Circuits and Systems for Video Technology}} \bibinfo{volume}{34}, \bibinfo{number}{2} (\bibinfo{year}{2024}), \bibinfo{pages}{1020--1031}.
\newblock
\href{https://doi.org/10.1109/TCSVT.2023.3289624}{doi:\nolinkurl{10.1109/TCSVT.2023.3289624}}


\bibitem[Huang et~al\mbox{.}(2019)]%
        {got10k}
\bibfield{author}{\bibinfo{person}{Lianghua Huang}, \bibinfo{person}{Xin Zhao}, {and} \bibinfo{person}{Kaiqi Huang}.} \bibinfo{year}{2019}\natexlab{}.
\newblock \showarticletitle{Got-10k: A large high-diversity benchmark for generic object tracking in the wild}.
\newblock \bibinfo{journal}{\emph{IEEE transactions on pattern analysis and machine intelligence}} \bibinfo{volume}{43}, \bibinfo{number}{5} (\bibinfo{year}{2019}), \bibinfo{pages}{1562--1577}.
\newblock


\bibitem[Li et~al\mbox{.}(2019)]%
        {siamrpn++}
\bibfield{author}{\bibinfo{person}{Bo Li}, \bibinfo{person}{Wei Wu}, \bibinfo{person}{Qiang Wang}, \bibinfo{person}{Fangyi Zhang}, \bibinfo{person}{Junliang Xing}, {and} \bibinfo{person}{Junjie Yan}.} \bibinfo{year}{2019}\natexlab{}.
\newblock \showarticletitle{Siamrpn++: Evolution of siamese visual tracking with very deep networks}. In \bibinfo{booktitle}{\emph{Proceedings of the IEEE/CVF conference on computer vision and pattern recognition}}. \bibinfo{pages}{4282--4291}.
\newblock


\bibitem[Li et~al\mbox{.}(2018)]%
        {siamrpn}
\bibfield{author}{\bibinfo{person}{Bo Li}, \bibinfo{person}{Junjie Yan}, \bibinfo{person}{Wei Wu}, \bibinfo{person}{Zheng Zhu}, {and} \bibinfo{person}{Xiaolin Hu}.} \bibinfo{year}{2018}\natexlab{}.
\newblock \showarticletitle{High performance visual tracking with siamese region proposal network}. In \bibinfo{booktitle}{\emph{Proceedings of the IEEE conference on computer vision and pattern recognition}}. \bibinfo{pages}{8971--8980}.
\newblock


\bibitem[Li et~al\mbox{.}(2024)]%
        {mambalct}
\bibfield{author}{\bibinfo{person}{Xiaohai Li}, \bibinfo{person}{Bineng Zhong}, \bibinfo{person}{Qihua Liang}, \bibinfo{person}{Guorong Li}, \bibinfo{person}{Zhiyi Mo}, {and} \bibinfo{person}{Shuxiang Song}.} \bibinfo{year}{2024}\natexlab{}.
\newblock \showarticletitle{MambaLCT: Boosting Tracking via Long-term Context State Space Model}.
\newblock \bibinfo{journal}{\emph{arXiv preprint arXiv:2412.13615}} (\bibinfo{year}{2024}).
\newblock


\bibitem[Lin(2017)]%
        {focal_loss}
\bibfield{author}{\bibinfo{person}{T Lin}.} \bibinfo{year}{2017}\natexlab{}.
\newblock \showarticletitle{Focal Loss for Dense Object Detection}.
\newblock \bibinfo{journal}{\emph{arXiv preprint arXiv:1708.02002}} (\bibinfo{year}{2017}).
\newblock


\bibitem[Lin et~al\mbox{.}(2014)]%
        {coco}
\bibfield{author}{\bibinfo{person}{Tsung-Yi Lin}, \bibinfo{person}{Michael Maire}, \bibinfo{person}{Serge Belongie}, \bibinfo{person}{James Hays}, \bibinfo{person}{Pietro Perona}, \bibinfo{person}{Deva Ramanan}, \bibinfo{person}{Piotr Doll{\'a}r}, {and} \bibinfo{person}{C~Lawrence Zitnick}.} \bibinfo{year}{2014}\natexlab{}.
\newblock \showarticletitle{Microsoft coco: Common objects in context}. In \bibinfo{booktitle}{\emph{Computer vision--ECCV 2014: 13th European conference, zurich, Switzerland, September 6-12, 2014, proceedings, part v 13}}. Springer, \bibinfo{pages}{740--755}.
\newblock


\bibitem[Liu et~al\mbox{.}(2024)]%
        {vmamba}
\bibfield{author}{\bibinfo{person}{Yue Liu}, \bibinfo{person}{Yunjie Tian}, \bibinfo{person}{Yuzhong Zhao}, \bibinfo{person}{Hongtian Yu}, \bibinfo{person}{Lingxi Xie}, \bibinfo{person}{Yaowei Wang}, \bibinfo{person}{Qixiang Ye}, \bibinfo{person}{Jianbin Jiao}, {and} \bibinfo{person}{Yunfan Liu}.} \bibinfo{year}{2024}\natexlab{}.
\newblock \bibinfo{title}{VMamba: Visual State Space Model}.
\newblock
\showeprint[arxiv]{2401.10166}~[cs.CV]
\urldef\tempurl%
\url{https://arxiv.org/abs/2401.10166}
\showURL{%
\tempurl}


\bibitem[Loshchilov and Hutter(2019)]%
        {AdamW}
\bibfield{author}{\bibinfo{person}{Ilya Loshchilov} {and} \bibinfo{person}{Frank Hutter}.} \bibinfo{year}{2019}\natexlab{}.
\newblock \bibinfo{title}{Decoupled Weight Decay Regularization}.
\newblock
\showeprint[arxiv]{1711.05101}~[cs.LG]
\urldef\tempurl%
\url{https://arxiv.org/abs/1711.05101}
\showURL{%
\tempurl}


\bibitem[Muller et~al\mbox{.}(2018)]%
        {TrackingNet}
\bibfield{author}{\bibinfo{person}{Matthias Muller}, \bibinfo{person}{Adel Bibi}, \bibinfo{person}{Silvio Giancola}, \bibinfo{person}{Salman Alsubaihi}, {and} \bibinfo{person}{Bernard Ghanem}.} \bibinfo{year}{2018}\natexlab{}.
\newblock \showarticletitle{Trackingnet: A large-scale dataset and benchmark for object tracking in the wild}. In \bibinfo{booktitle}{\emph{Proceedings of the European conference on computer vision (ECCV)}}. \bibinfo{pages}{300--317}.
\newblock


\bibitem[Rezatofighi et~al\mbox{.}(2019)]%
        {giou_loss}
\bibfield{author}{\bibinfo{person}{Hamid Rezatofighi}, \bibinfo{person}{Nathan Tsoi}, \bibinfo{person}{JunYoung Gwak}, \bibinfo{person}{Amir Sadeghian}, \bibinfo{person}{Ian Reid}, {and} \bibinfo{person}{Silvio Savarese}.} \bibinfo{year}{2019}\natexlab{}.
\newblock \showarticletitle{Generalized intersection over union: A metric and a loss for bounding box regression}. In \bibinfo{booktitle}{\emph{Proceedings of the IEEE/CVF conference on computer vision and pattern recognition}}. \bibinfo{pages}{658--666}.
\newblock


\bibitem[Shi et~al\mbox{.}(2025)]%
        {MAVLT}
\bibfield{author}{\bibinfo{person}{Liangtao Shi}, \bibinfo{person}{Bineng Zhong}, \bibinfo{person}{Qihua Liang}, \bibinfo{person}{Xiantao Hu}, \bibinfo{person}{Zhiyi Mo}, {and} \bibinfo{person}{Shuxiang Song}.} \bibinfo{year}{2025}\natexlab{}.
\newblock \showarticletitle{Mamba Adapter: Efficient Multi-Modal Fusion for Vision-Language Tracking}.
\newblock \bibinfo{journal}{\emph{IEEE Transactions on Circuits and Systems for Video Technology}} (\bibinfo{year}{2025}), \bibinfo{pages}{1--1}.
\newblock
\href{https://doi.org/10.1109/TCSVT.2025.3557570}{doi:\nolinkurl{10.1109/TCSVT.2025.3557570}}


\bibitem[Shi et~al\mbox{.}(2024)]%
        {evptrack}
\bibfield{author}{\bibinfo{person}{Liangtao Shi}, \bibinfo{person}{Bineng Zhong}, \bibinfo{person}{Qihua Liang}, \bibinfo{person}{Ning Li}, \bibinfo{person}{Shengping Zhang}, {and} \bibinfo{person}{Xianxian Li}.} \bibinfo{year}{2024}\natexlab{}.
\newblock \showarticletitle{Explicit Visual Prompts for Visual Object Tracking}. In \bibinfo{booktitle}{\emph{Proceedings of the AAAI Conference on Artificial Intelligence}}, Vol.~\bibinfo{volume}{38}. \bibinfo{pages}{4838--4846}.
\newblock


\bibitem[Smith et~al\mbox{.}(2023)]%
        {ssm3}
\bibfield{author}{\bibinfo{person}{Jimmy T.~H. Smith}, \bibinfo{person}{Shalini~De Mello}, \bibinfo{person}{Jan Kautz}, \bibinfo{person}{Scott~W. Linderman}, {and} \bibinfo{person}{Wonmin Byeon}.} \bibinfo{year}{2023}\natexlab{}.
\newblock \bibinfo{title}{Convolutional State Space Models for Long-Range Spatiotemporal Modeling}.
\newblock
\showeprint[arxiv]{2310.19694}~[cs.LG]
\urldef\tempurl%
\url{https://arxiv.org/abs/2310.19694}
\showURL{%
\tempurl}


\bibitem[Tang et~al\mbox{.}(2023)]%
        {DFAT}
\bibfield{author}{\bibinfo{person}{Zhangyong Tang}, \bibinfo{person}{Tianyang Xu}, \bibinfo{person}{Hui Li}, \bibinfo{person}{Xiao-Jun Wu}, \bibinfo{person}{Xuefeng Zhu}, {and} \bibinfo{person}{Josef Kittler}.} \bibinfo{year}{2023}\natexlab{}.
\newblock \showarticletitle{Exploring fusion strategies for accurate RGBT visual object tracking}.
\newblock \bibinfo{journal}{\emph{Information Fusion}}  \bibinfo{volume}{99} (\bibinfo{year}{2023}), \bibinfo{pages}{101881}.
\newblock


\bibitem[Tang et~al\mbox{.}(2024)]%
        {GMMT}
\bibfield{author}{\bibinfo{person}{Zhangyong Tang}, \bibinfo{person}{Tianyang Xu}, \bibinfo{person}{Xiaojun Wu}, \bibinfo{person}{Xue-Feng Zhu}, {and} \bibinfo{person}{Josef Kittler}.} \bibinfo{year}{2024}\natexlab{}.
\newblock \showarticletitle{Generative-based fusion mechanism for multi-modal tracking}. In \bibinfo{booktitle}{\emph{Proceedings of the AAAI Conference on Artificial Intelligence}}, Vol.~\bibinfo{volume}{38}. \bibinfo{pages}{5189--5197}.
\newblock


\bibitem[Tian et~al\mbox{.}(2024)]%
        {Fast-iTPN}
\bibfield{author}{\bibinfo{person}{Yunjie Tian}, \bibinfo{person}{Lingxi Xie}, \bibinfo{person}{Jihao Qiu}, \bibinfo{person}{Jianbin Jiao}, \bibinfo{person}{Yaowei Wang}, \bibinfo{person}{Qi Tian}, {and} \bibinfo{person}{Qixiang Ye}.} \bibinfo{year}{2024}\natexlab{}.
\newblock \showarticletitle{Fast-iTPN: Integrally pre-trained transformer pyramid network with token migration}.
\newblock \bibinfo{journal}{\emph{IEEE Transactions on Pattern Analysis and Machine Intelligence}} (\bibinfo{year}{2024}).
\newblock


\bibitem[Wang et~al\mbox{.}(2021)]%
        {tnl2k}
\bibfield{author}{\bibinfo{person}{Xiao Wang}, \bibinfo{person}{Xiujun Shu}, \bibinfo{person}{Zhipeng Zhang}, \bibinfo{person}{Bo Jiang}, \bibinfo{person}{Yaowei Wang}, \bibinfo{person}{Yonghong Tian}, {and} \bibinfo{person}{Feng Wu}.} \bibinfo{year}{2021}\natexlab{}.
\newblock \showarticletitle{Towards more flexible and accurate object tracking with natural language: Algorithms and benchmark}. In \bibinfo{booktitle}{\emph{Proceedings of the IEEE/CVF conference on computer vision and pattern recognition}}. \bibinfo{pages}{13763--13773}.
\newblock


\bibitem[Wei et~al\mbox{.}(2023)]%
        {artrack}
\bibfield{author}{\bibinfo{person}{Xing Wei}, \bibinfo{person}{Yifan Bai}, \bibinfo{person}{Yongchao Zheng}, \bibinfo{person}{Dahu Shi}, {and} \bibinfo{person}{Yihong Gong}.} \bibinfo{year}{2023}\natexlab{}.
\newblock \showarticletitle{Autoregressive visual tracking}. In \bibinfo{booktitle}{\emph{Proceedings of the IEEE/CVF Conference on Computer Vision and Pattern Recognition}}. \bibinfo{pages}{9697--9706}.
\newblock


\bibitem[Woo et~al\mbox{.}(2018)]%
        {ca_sa}
\bibfield{author}{\bibinfo{person}{Sanghyun Woo}, \bibinfo{person}{Jongchan Park}, \bibinfo{person}{Joon-Young Lee}, {and} \bibinfo{person}{In~So Kweon}.} \bibinfo{year}{2018}\natexlab{}.
\newblock \bibinfo{title}{CBAM: Convolutional Block Attention Module}.
\newblock
\showeprint[arxiv]{1807.06521}~[cs.CV]
\urldef\tempurl%
\url{https://arxiv.org/abs/1807.06521}
\showURL{%
\tempurl}


\bibitem[Xie et~al\mbox{.}(2022)]%
        {sbt}
\bibfield{author}{\bibinfo{person}{Fei Xie}, \bibinfo{person}{Chunyu Wang}, \bibinfo{person}{Guangting Wang}, \bibinfo{person}{Yue Cao}, \bibinfo{person}{Wankou Yang}, {and} \bibinfo{person}{Wenjun Zeng}.} \bibinfo{year}{2022}\natexlab{}.
\newblock \bibinfo{title}{Correlation-Aware Deep Tracking}.
\newblock
\showeprint[arxiv]{2203.01666}~[cs.CV]
\urldef\tempurl%
\url{https://arxiv.org/abs/2203.01666}
\showURL{%
\tempurl}


\bibitem[Xie et~al\mbox{.}(2024a)]%
        {temtrack}
\bibfield{author}{\bibinfo{person}{Jinxia Xie}, \bibinfo{person}{Bineng Zhong}, \bibinfo{person}{Qihua Liang}, \bibinfo{person}{Ning Li}, \bibinfo{person}{Zhiyi Mo}, {and} \bibinfo{person}{Shuxiang Song}.} \bibinfo{year}{2024}\natexlab{a}.
\newblock \showarticletitle{Robust Tracking via Mamba-based Context-aware Token Learning}.
\newblock \bibinfo{journal}{\emph{arXiv preprint arXiv:2412.13611}} (\bibinfo{year}{2024}).
\newblock


\bibitem[Xie et~al\mbox{.}(2024b)]%
        {aqatrack}
\bibfield{author}{\bibinfo{person}{Jinxia Xie}, \bibinfo{person}{Bineng Zhong}, \bibinfo{person}{Zhiyi Mo}, \bibinfo{person}{Shengping Zhang}, \bibinfo{person}{Liangtao Shi}, \bibinfo{person}{Shuxiang Song}, {and} \bibinfo{person}{Rongrong Ji}.} \bibinfo{year}{2024}\natexlab{b}.
\newblock \showarticletitle{Autoregressive Queries for Adaptive Tracking with Spatio-Temporal Transformers}. In \bibinfo{booktitle}{\emph{Proceedings of the IEEE/CVF Conference on Computer Vision and Pattern Recognition}}. \bibinfo{pages}{19300--19309}.
\newblock


\bibitem[Xu et~al\mbox{.}(2020)]%
        {siamfc++}
\bibfield{author}{\bibinfo{person}{Yinda Xu}, \bibinfo{person}{Zeyu Wang}, \bibinfo{person}{Zuoxin Li}, \bibinfo{person}{Ye Yuan}, {and} \bibinfo{person}{Gang Yu}.} \bibinfo{year}{2020}\natexlab{}.
\newblock \bibinfo{title}{SiamFC++: Towards Robust and Accurate Visual Tracking with Target Estimation Guidelines}.
\newblock
\showeprint[arxiv]{1911.06188}~[cs.CV]
\urldef\tempurl%
\url{https://arxiv.org/abs/1911.06188}
\showURL{%
\tempurl}


\bibitem[Xue et~al\mbox{.}(2024)]%
        {uscltrack}
\bibfield{author}{\bibinfo{person}{Chaocan Xue}, \bibinfo{person}{Bineng Zhong}, \bibinfo{person}{Qihua Liang}, \bibinfo{person}{Haiying Xia}, {and} \bibinfo{person}{Shuxiang Song}.} \bibinfo{year}{2024}\natexlab{}.
\newblock \showarticletitle{Unifying Motion and Appearance Cues for Visual Tracking via Shared Queries}.
\newblock \bibinfo{journal}{\emph{IEEE Transactions on Circuits and Systems for Video Technology}} (\bibinfo{year}{2024}).
\newblock


\bibitem[Xue et~al\mbox{.}(2025)]%
        {SGLATrack}
\bibfield{author}{\bibinfo{person}{Chaocan Xue}, \bibinfo{person}{Bineng Zhong}, \bibinfo{person}{Qihua Liang}, \bibinfo{person}{Yaozong Zheng}, \bibinfo{person}{Ning Li}, \bibinfo{person}{Yuanliang Xue}, {and} \bibinfo{person}{Shuxiang Song}.} \bibinfo{year}{2025}\natexlab{}.
\newblock \bibinfo{title}{Similarity-Guided Layer-Adaptive Vision Transformer for UAV Tracking}.
\newblock
\showeprint[arxiv]{2503.06625}~[cs.CV]
\urldef\tempurl%
\url{https://arxiv.org/abs/2503.06625}
\showURL{%
\tempurl}


\bibitem[Yan et~al\mbox{.}(2021)]%
        {stark}
\bibfield{author}{\bibinfo{person}{Bin Yan}, \bibinfo{person}{Houwen Peng}, \bibinfo{person}{Jianlong Fu}, \bibinfo{person}{Dong Wang}, {and} \bibinfo{person}{Huchuan Lu}.} \bibinfo{year}{2021}\natexlab{}.
\newblock \showarticletitle{Learning spatio-temporal transformer for visual tracking}. In \bibinfo{booktitle}{\emph{Proceedings of the IEEE/CVF international conference on computer vision}}. \bibinfo{pages}{10448--10457}.
\newblock


\bibitem[Yang et~al\mbox{.}(2023)]%
        {F-BDMTrack}
\bibfield{author}{\bibinfo{person}{Dawei Yang}, \bibinfo{person}{Jianfeng He}, \bibinfo{person}{Yinchao Ma}, \bibinfo{person}{Qianjin Yu}, {and} \bibinfo{person}{Tianzhu Zhang}.} \bibinfo{year}{2023}\natexlab{}.
\newblock \showarticletitle{Foreground-Background Distribution Modeling Transformer for Visual Object Tracking}. In \bibinfo{booktitle}{\emph{Proceedings of the IEEE/CVF International Conference on Computer Vision (ICCV)}}. \bibinfo{pages}{10117--10127}.
\newblock


\bibitem[Yang et~al\mbox{.}(2024)]%
        {vivim}
\bibfield{author}{\bibinfo{person}{Yijun Yang}, \bibinfo{person}{Zhaohu Xing}, \bibinfo{person}{Lequan Yu}, \bibinfo{person}{Chunwang Huang}, \bibinfo{person}{Huazhu Fu}, {and} \bibinfo{person}{Lei Zhu}.} \bibinfo{year}{2024}\natexlab{}.
\newblock \bibinfo{title}{Vivim: a Video Vision Mamba for Medical Video Segmentation}.
\newblock
\showeprint[arxiv]{2401.14168}~[cs.CV]
\urldef\tempurl%
\url{https://arxiv.org/abs/2401.14168}
\showURL{%
\tempurl}


\bibitem[Ye et~al\mbox{.}(2022)]%
        {ostrack}
\bibfield{author}{\bibinfo{person}{Botao Ye}, \bibinfo{person}{Hong Chang}, \bibinfo{person}{Bingpeng Ma}, \bibinfo{person}{Shiguang Shan}, {and} \bibinfo{person}{Xilin Chen}.} \bibinfo{year}{2022}\natexlab{}.
\newblock \showarticletitle{Joint feature learning and relation modeling for tracking: A one-stream framework}. In \bibinfo{booktitle}{\emph{European Conference on Computer Vision}}. Springer, \bibinfo{pages}{341--357}.
\newblock


\bibitem[Zhang et~al\mbox{.}(2020)]%
        {Ocean}
\bibfield{author}{\bibinfo{person}{Zhipeng Zhang}, \bibinfo{person}{Houwen Peng}, \bibinfo{person}{Jianlong Fu}, \bibinfo{person}{Bing Li}, {and} \bibinfo{person}{Weiming Hu}.} \bibinfo{year}{2020}\natexlab{}.
\newblock \showarticletitle{Ocean: Object-aware anchor-free tracking}. In \bibinfo{booktitle}{\emph{European conference on computer vision}}. Springer, \bibinfo{pages}{771--787}.
\newblock


\bibitem[Zheng et~al\mbox{.}(2024)]%
        {odtrack}
\bibfield{author}{\bibinfo{person}{Yaozong Zheng}, \bibinfo{person}{Bineng Zhong}, \bibinfo{person}{Qihua Liang}, \bibinfo{person}{Zhiyi Mo}, \bibinfo{person}{Shengping Zhang}, {and} \bibinfo{person}{Xianxian Li}.} \bibinfo{year}{2024}\natexlab{}.
\newblock \showarticletitle{Odtrack: Online dense temporal token learning for visual tracking}. In \bibinfo{booktitle}{\emph{Proceedings of the AAAI Conference on Artificial Intelligence}}, Vol.~\bibinfo{volume}{38}. \bibinfo{pages}{7588--7596}.
\newblock


\bibitem[Zhu et~al\mbox{.}(2024)]%
        {vim}
\bibfield{author}{\bibinfo{person}{Lianghui Zhu}, \bibinfo{person}{Bencheng Liao}, \bibinfo{person}{Qian Zhang}, \bibinfo{person}{Xinlong Wang}, \bibinfo{person}{Wenyu Liu}, {and} \bibinfo{person}{Xinggang Wang}.} \bibinfo{year}{2024}\natexlab{}.
\newblock \bibinfo{title}{Vision Mamba: Efficient Visual Representation Learning with Bidirectional State Space Model}.
\newblock
\showeprint[arxiv]{2401.09417}~[cs.CV]
\urldef\tempurl%
\url{https://arxiv.org/abs/2401.09417}
\showURL{%
\tempurl}


\end{thebibliography}
\end{document}